\PassOptionsToPackage{table}{xcolor}
\documentclass[10pt,a4paper,logo]{googledeepmind}
\usepackage{times}
\usepackage{natbib}
\setcitestyle{authoryear,round,citesep={;},aysep={,},yysep={;}}
\newcommand{\paperCodeURL}{https://github.com/Eureka-Maggie/RARE}

\usepackage{amsmath,amsfonts,bm}

\def\eqref#1{equation~\ref{#1}}

\def\1{\bm{1}}

\DeclareMathAlphabet{\mathsfit}{\encodingdefault}{\sfdefault}{m}{sl}
\SetMathAlphabet{\mathsfit}{bold}{\encodingdefault}{\sfdefault}{bx}{n}

\usepackage{amsmath}
\usepackage{amssymb}
\usepackage{array}
\usepackage{booktabs}
\usepackage{colortbl}
\usepackage{float}
\usepackage{graphicx}
\usepackage{hyperref}
\usepackage{listings}
\usepackage{longtable}
\usepackage{multirow}
\usepackage{placeins}
\usepackage{rotating}
\usepackage{tabularx}
\usepackage{tcolorbox}
\usepackage{url}
\usepackage{wrapfig}
\usepackage{xspace}

\newcommand{\ours}{\textup{\textsc{RARE}}\xspace}

\definecolor{morandiBlue}{RGB}{232,244,250}
\definecolor{morandiSage}{RGB}{237,247,234}
\definecolor{morandiRose}{RGB}{252,238,241}
\definecolor{morandiSand}{RGB}{252,247,226}

\newtcolorbox{promptbox}[1]{
  enhanced,
  breakable,
  colback=black!3,
  colframe=black!20,
  colbacktitle=black!9,
  coltitle=black,
  fonttitle=\bfseries\small,
  before upper={\lstset{aboveskip=0pt,belowskip=0pt}},
  title={#1},
  boxrule=0.5pt,
  arc=1.5mm,
  left=2.5mm,
  right=2.5mm,
  top=1.5mm,
  bottom=1.5mm
}

\title{\ours{}: Reinforcement Learning with Adaptive Rubric Evolution for Open-Ended Generation}

\input{authors_bu}

\begin{abstract}
Verifiable rewards have advanced reinforcement learning in mathematics and code.
However, many valuable open-ended tasks share three properties: no automatic verifier of overall quality, multiple requirements to satisfy jointly, and multiple valid responses to each query.
Training therefore needs feedback on requirement satisfaction across different valid responses.
Rubric-based reinforcement learning provides this feedback by aggregating scores from explicit criteria.
As learning progresses unevenly, some criteria become fully satisfied and lose discrimination while others remain informative.
Dynamic methods revise shared rubrics or select evolving criteria to restore discrimination, but may also change useful criteria.
To address this, we propose \textbf{R}einforcement Learning with \textbf{A}daptive \textbf{R}ubric \textbf{E}volution (\ours{}).
The core idea is to use current policy feedback to guide both which criteria change and what their replacements assess.
Sample-level updates replace criteria fully satisfied across current outputs with checks of remaining quality differences, retaining all other criteria and scores.
Within this rule, we compare rewarding desired qualities with penalizing observed flaws.
Video script generation exemplifies these properties and supports content production.
Its demands for coherent, executable narratives under user constraints make it a challenging testbed.
Targeting flaws performs best across three video types with a mean gain of 3.82 points over matched static rubrics.
Model and human preferences support these gains.
Improvements also extend to writing and general generation benchmarks.

\end{abstract}

\begin{document}

\maketitle

\section{Introduction}
\label{sec:introduction}
Reinforcement learning with verifiable rewards (RLVR) trains models for mathematical reasoning and code generation using automatic correctness checks \citep{shao2024deepseekmath,guo2025deepseekr1}. Many valuable open-ended tasks, however, lack a verifier of overall quality \citep{gunjal2026rubrics}. Their outputs must meet several requirements jointly, while each query admits multiple valid responses \citep{wu2025writingbench}. Training feedback must therefore assess requirements across valid responses.

Rubric-based reinforcement learning supplies this feedback by scoring outputs against explicit criteria \citep{gunjal2026rubrics,wei2026qurl}. During training, progress across requirements can be uneven. A criterion becomes saturated when every sampled output receives full credit, so it contributes no reward difference within the group \citep{shao2024deepseekmath}. Other criteria may still distinguish quality. This partial saturation motivates adapting the rubric as the policy improves.

Dynamic methods address this need by revising shared rubrics \citep{kim2025evaluative} or generating criteria from response comparisons \citep{rezaei2025online}. Some preserve initial requirements while selecting evolving criteria by score variation \citep{shao2025drtulu}. Yet preserving initial requirements does not determine which later criteria should remain unchanged. Such criteria may still distinguish current outputs, so revision and pool selection can remove useful feedback. The challenge is therefore to revise criteria for remaining weaknesses while preserving useful feedback.

To address this challenge, we propose \textbf{R}einforcement Learning with \textbf{A}daptive \textbf{R}ubric \textbf{E}volution (\ours{}). Our core idea is to guide rubric revision by the policy's uneven progress across requirements. At sample level, current scores determine which criteria may change, while response differences guide what new criteria assess. The policy update retains all other criteria and their scores.

Video script generation supports content production and admits varied narratives for each request \citep{mirowski2023cowriting}. Producing coherent, executable scripts requires satisfying user constraints while maintaining character consistency and causal logic across scenes \citep{zheng2025cml}. Even scripts covering all requested elements may differ in these qualities. These demands make script generation a challenging testbed for selective rubric revision.

Concretely, criteria become eligible for replacement when all current outputs receive full credit. The generator proposes criteria for quality differences absent from the retained rubric, and the judge rescores them on those outputs. We compare desired qualities (\textit{Positive}), observed flaws (\textit{Negative}), and both (\textit{Mixed}) as replacement targets. At task level, we periodically revise each video type's shared rubric using representative outputs across queries and reward levels.

We construct 5,989 query--script pairs across three video types and evaluate on 596 held-out queries using twelve fixed metrics. With Qwen3-4B-Instruct \citep{qwen2025qwen3model}, all four \ours{} variants outperform matched static and adapted dynamic baselines. Sample-level \textit{Negative} performs best with a mean gain of 3.82 points over fixed per-query rubrics and improves all four quality categories. Task-level revision gains 1.08 points over fixed shared rubrics. Both figures are averages across video types. Blind model and human preferences support \textit{Negative}'s gains. Analyses show that replacements distinguish current and newly sampled scripts while earlier requirements remain largely satisfied. Improvements over static training also extend to WritingBench \citep{wu2025writingbench} and BiGGen Bench \citep{kim2025biggen}.

Our contributions are summarized below.
\begin{itemize}
    \item \textbf{Rubric evolution guided by policy feedback.} We develop \ours{} to adapt rewards to uneven learning progress. Sample-level updates combine selective replacement with criteria grounded in response differences, preserving all other criteria and scores.
    \item \textbf{Insights into effective rubric evolution.} Sample-level \textit{Negative} yields the best script quality across the studied update levels and replacement directions. Among the three directions, it most often restores score variation. Its new criteria distinguish later scripts while earlier requirements remain largely satisfied.
    \item \textbf{Improved generation quality.} We train policies separately for each video type and text task. \ours{} outperforms matched static and adapted dynamic baselines on three video types. It also outperforms static baselines on two additional open-ended text tasks.
\end{itemize}

\suppressfloats[t]
\begin{figure}[t]
    \centering
    \includegraphics[width=\linewidth]{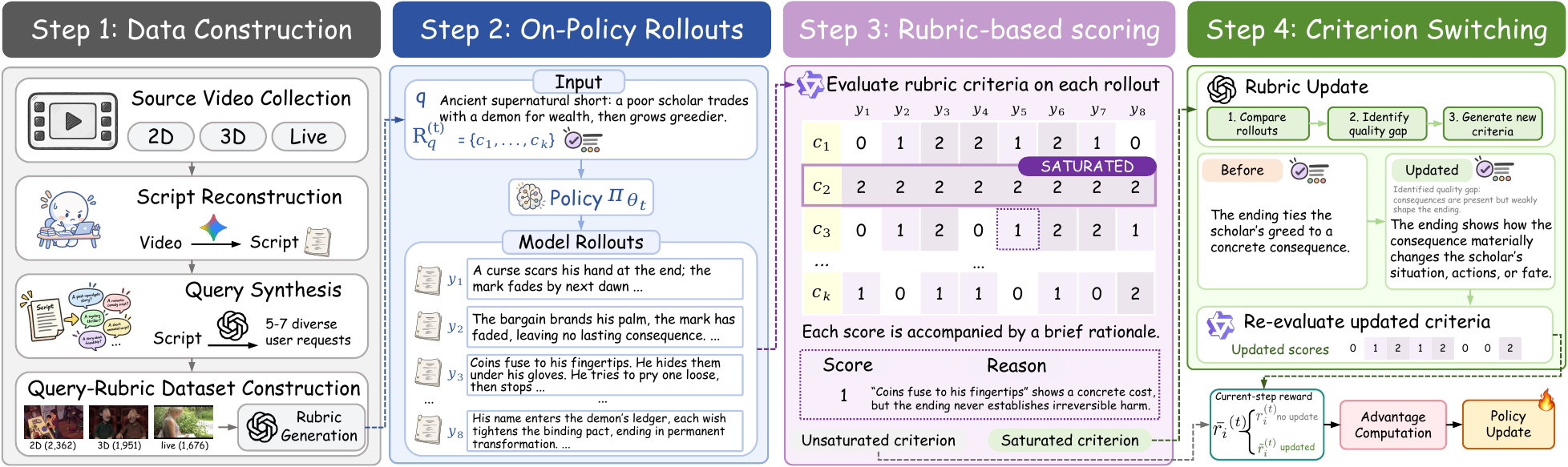}
    \caption{\textbf{Sample-level \ours{} replaces saturated criteria while retaining other feedback.} Illustrated on script generation, the generator proposes replacements from rollout differences. The judge rescores the replacements on the same responses before the policy update.}
    \label{fig:training-pipeline}
\end{figure}

\section{Related Work}
\label{sec:related-work}
\subsection{Rubric-Based Rewards for Open-Ended Generation}

Rubric-based rewards support RL for open-ended text generation when overall quality cannot be checked by an exact verifier \citep{gunjal2026rubrics,wei2026qurl}.
They decompose quality into explicit criteria written by humans or synthesized by language models \citep{huang2025anchors}.
Construction methods extract checklists from instructions \citep{viswanathan2025checklists} or derive criteria by contrasting preferred and rejected responses \citep{liu2026openrubrics}.
Filtering removes rubrics that misrank these pairs.
Judges score the retained criteria to provide feedback on individual quality requirements.
However, fixed criteria can lose discrimination as outputs change, and optimizing incomplete criteria can permit reward hacking \citep{mahmoud2026rewardhacking}.
\subsection{Evolving Rubric Rewards}

Task-level methods periodically revise shared rubrics using recent policy outputs \citep{kim2025evaluative}.
Sample-level methods update query-specific criteria from response differences \citep{rezaei2025online,shao2025drtulu}.
Updates use an external model or a trained rubric generator \citep{guan2026evorubric,ding2026evorubrics}.
Deduplication and score-variance filtering select active criteria \citep{rezaei2025online,shao2025drtulu}, while memory and synthetic comparisons guide refinement \citep{wu2026amaris,yang2026rubricstrial}.
Reinforcement Learning with Evolving Rubrics (RLER) preserves initial requirements alongside evolving criteria.
Our sample-level rule restricts replacement to saturated criteria and retains all other current criteria and scores for the policy update.
\subsection{Script Generation as an Open-Ended Task}

Script generation uses hierarchical planning to develop a brief premise into a complete screenplay \citep{mirowski2023cowriting}, while agent collaboration supports screenwriting and interactive drama \citep{chen2024hollmwood,han2024ibsen}.
Other methods learn from scene-level supervision \citep{gandhi2023kurosawa} or convert dialogue and multimodal materials into cinematic and advertising scripts \citep{mu2026script,hu2026mcsc}.
These approaches differ in input and generation process, but share the need to coordinate narrative content with scene-level detail.

Evaluation ranges from professional judgments of complete scripts \citep{mirowski2023cowriting} to tests of excerpts and continuations \citep{zheng2025cml,ma2025dramabench}.
Video-production protocols also evaluate scripts alongside rendered videos \citep{mu2026script,hu2026mcsc}.
Their measures vary with the task but broadly assess input adherence and narrative coherence alongside dramatic and cinematic effectiveness.
These requirements make script generation a useful setting for comparing evolving training feedback with fixed evaluation of request fulfillment and narrative quality.

\section{Method}
\label{sec:method}
\ours{} adapts rubric feedback to uneven learning progress using current outputs and criterion scores.
Sample-level updates replace saturated criteria with checks of remaining quality differences while retaining all other criteria and scores.
Task-level updates use representative outputs across queries to guide periodic revisions of shared rubrics.

\subsection{Rubric Rewards and Policy Optimization}

Given a text query $q$, the policy $\pi_\theta(\cdot\mid q)$ generates a response.
At training step $t$, we sample $G$ responses
$y_i\sim\pi_{\theta_t}(\cdot\mid q)$ for $i\in\{1,\ldots,G\}$.
The judge evaluates them using the assigned rubric
$\mathcal{R}_q^{(t)}=\{c_1^{(t)},\ldots,c_k^{(t)}\}$.
This rubric determines rewards and is separate from the policy input.

Each criterion $c_j^{(t)}$ defines an integer scoring rule with maximum $m_j$.
The judge returns a score $s_{i,j}^{(t)}\in\{0,\ldots,m_j\}$ and a rationale for each response.
Their normalized sum gives the reward
$r_i^{(t)}=(\sum_{j=1}^{k}s_{i,j}^{(t)})/(\sum_{j=1}^{k}m_j)$.
Revisions preserve criterion count, identifiers, and maxima, so the denominator and reward range $[0,1]$ remain fixed throughout training.

Sample-level replacements are rescored on the same responses before optimization.
The update uses $\bar r_i^{(t)}=\tilde r_i^{(t)}$ after an accepted revision (Eq.~\ref{eq:updated-reward}) and $\bar r_i^{(t)}=r_i^{(t)}$ otherwise.
Group Relative Policy Optimization (GRPO) \citep{shao2024deepseekmath} uses advantages $\hat A_i=(\bar r_i^{(t)}-\mu_t)/(\sigma_t+\epsilon)$, with $\epsilon>0$ for numerical stability.
Here $\mu_t$ and $\sigma_t$ denote the group reward mean and standard deviation.
We use a token-level clipped objective with Kullback--Leibler (KL) regularization.
The optimization procedure remains unchanged as rubric revisions modify the training rewards.

\subsection{Sample-Level Criterion Replacement}

We identify full-score saturation in the current group through the eligible set
\begin{equation}
    \mathcal{S}_q^{(t)}
    =
    \left\{
        j\in\{1,\ldots,k\}
        \mid
        s_{i,j}^{(t)}=m_j
        \ \forall i\in\{1,\ldots,G\}
    \right\}.
    \label{eq:saturated-set}
\end{equation}
Each eligible criterion contributes a constant that cancels in group-relative centering.
Uniformly low scores can indicate an unmet requirement, so those criteria remain in the rubric.

When $\mathcal{S}_q^{(t)}$ is nonempty, we request
$b_t=\min(B,|\mathcal{S}_q^{(t)}|)$ replacements.
Saturation determines eligibility, while the budget $B$ controls the extent of each revision.
The generator selects $\mathcal{U}_q^{(t)}\subseteq\mathcal{S}_q^{(t)}$ with $|\mathcal{U}_q^{(t)}|=b_t$, leaving all other criteria and scores fixed.
This confines reward changes to the selected criteria and preserves the contribution of retained scores to the current update.
Appendix~\ref{app:trigger-sensitivity} examines the budget and group size using observed revision events and score matrices.

The generator compares current responses using the query, rubric, and judge feedback (Figure~\ref{fig:training-pipeline}).
Quality differences not captured by retained criteria guide each replacement $c_j^{\prime(t)}$.
Every replacement preserves its identifier and maximum $m_j$ and defines all integer score levels.
Criteria assess observable qualities without prescribing a response, accommodating different valid realizations.

We compare three directions for generating replacement criteria.
\textit{Positive} gives full credit when a desired quality is fully realized.
\textit{Negative} gives full credit when an observed flaw is absent and lowers scores with its severity.
This directs learning toward reducing a specific defect while leaving its remedy open to different valid realizations of the request.
\textit{Mixed} permits both directions and requires both when at least two criteria are replaced.
The variants share the eligibility rule, replacement budget, and scoring procedure, allowing us to compare these directions during training.

The judge jointly rescores replacements on the same $G$ responses, keeping the evidence fixed and providing feedback for the current update without additional sampling.
We combine the replacement scores $s_{i,j}^{\prime(t)}$ with the retained scores to compute
\begin{equation}
    \tilde r_i^{(t)}
    =
    \frac{
        \sum_{j\in\mathcal{U}_q^{(t)}}s_{i,j}^{\prime(t)}
        +
        \sum_{j\notin\mathcal{U}_q^{(t)}}s_{i,j}^{(t)}
    }{
        \sum_{j=1}^{k}m_j
    }.
    \label{eq:updated-reward}
\end{equation}
Validated revisions supply $\tilde r_i^{(t)}$ for the current update and persist across visits to $q$.
Otherwise, the original rubric and rewards remain in use.
Appendix~\ref{app:sample-updates} details validation and persistence.

\subsection{Task-Level Rubric Evolution}

A shared rubric evaluates different queries, so revisions need evidence beyond one rollout group.
Our task-level extension periodically selects representatives across queries and group mean rewards, exposing different degrees of criterion satisfaction.
The generator strengthens easily satisfied checks and reformulates overly difficult criteria as observable prerequisites.
Criterion count, identifiers, maxima, and reward aggregation remain fixed across task-level revisions.

Rescoring the same representatives tests discrimination under fixed evidence.
Acceptance requires at least two distinct rewards and a bounded change in mean reward to limit shifts in reward scale.
Accepted revisions apply at the next step (Appendix~\ref{app:task-updates}).

\section{Experiments}
\label{sec:experiments}
We evaluate \ours{} primarily through video script generation, where explicit request constraints coexist with narrative and cinematic requirements.
A creative request permits different plots and scene structures, each of which must satisfy the request and develop a coherent narrative.
These properties make overall quality difficult to assess through exact output matching.
This setting lets us examine how rubric revisions change feedback and whether those changes accompany better outputs.
We compare update rules and replacement directions here.
Section~\ref{sec:generalization} then tests the sample-level procedure on additional text generation tasks using their own training data and evaluation criteria.

\subsection{Experimental Setup}

\subsubsection{Dataset}

\begin{wraptable}[9]{r}{0.47\linewidth}
    \centering
    \scriptsize
    \caption{\textbf{Dataset split by source video.}
    Entries show videos / queries.}
    \label{tab:dataset-statistics}
    \setlength{\tabcolsep}{2.7pt}
    \begin{tabular}{lccc}
        \toprule
        Video type & All & Train & Test \\
        \midrule
        2D animation & 370 / 2,362 & 333 / 2,125 & 37 / 237 \\
        3D/stop-motion & 302 / 1,951 & 272 / 1,759 & 30 / 192 \\
        Live action & 258 / 1,676 & 232 / 1,509 & 26 / 167 \\
        \midrule
        Total & \textbf{930 / 5,989} & \textbf{837 / 5,393} & \textbf{93 / 596} \\
        \bottomrule
    \end{tabular}
\end{wraptable}

Source videos come from the media collection used by AVA-Encoder \citep{li2026ava} and cover 2D animation, 3D and stop-motion animation, and live action.
Gemini 3.1 Pro \citep{google2026gemini} reconstructs a complete script from each video.
GPT-5.4 \citep{openai2026gpt54} then generates five to seven distinct creative queries from the script.
Each query is paired with the reconstructed script.
The queries express pre-production intent and omit endings, character names, and timestamps.

We split 5,989 pairs by source video into 5,393 training pairs and 596 test queries.
We train and evaluate each type separately to reflect scripting requirements for stylized 2D action, 3D staging, and live-action performance.
Table~\ref{tab:dataset-statistics} gives the split and Appendix~\ref{app:data-construction} details construction.

For each training query, GPT-5.4 constructs a rubric from scripts sampled from the initial policy.
We summarize these rubrics into a shared rubric for each video type.
The static and evolving variants thus start with the same requirements at each update level.
Appendix~\ref{app:data-construction} details rubric construction.

\subsubsection{Baselines and Variants}

All task-specific training starts from Qwen3-4B-Instruct-2507 \citep{yang2025qwen3}.
\textbf{Base} denotes the initial checkpoint without task-specific training.
\textbf{SFT} is trained on paired queries and reconstructed scripts.
All RL methods start directly from Base and share the training data, GRPO configuration, and generation settings.
We report mean RL scores over four training seeds.

\textbf{Sample-Static} uses a fixed rubric for each query.
\textbf{Task-Static} shares a fixed rubric within each video type.
We compare each with its corresponding \ours{} variant to measure the benefit of evolution at that level.
The sample-level \textit{Negative}, \textit{Positive}, and \textit{Mixed} variants share the trigger, budget, and scoring procedure to compare replacement directions.
The task-level variant uses evidence across queries to revise the rubric shared within each video type.

We adapt dynamic rubric methods at both update levels to the same script generation setting.
At task level, Meta Policy Optimization (MPO) \citep{kim2025evaluative} analyzes scored outputs and merges criterion refinements into a shared rubric.
Related methods vary scoring, memory use, and update frequency within this process \citep{wu2026amaris}.
At sample level, RLER \citep{shao2025drtulu} generates criteria from response differences and manages its evolving criterion pool by within-group score variation.
RLER retains initial requirements alongside the selected evolving criteria.
These baselines cover shared rubric revision and per-query criterion selection.

For script training, Qwen3.7-Plus \citep{alibaba2026qwen37plus} scores outputs and GPT-5.4 generates rubric revisions.
We additionally evaluate seven reference models under the same test protocol.
Training configurations and prompts appear in Appendices~\ref{app:training} and~\ref{app:prompts}.

\subsubsection{Evaluation Protocol}

Existing benchmarks cover scenes and continuations or scripts conditioned on dialogue and multimodal inputs \citep{gandhi2023kurosawa,zheng2025cml,ma2025dramabench,mu2026script,hu2026mcsc}.
To evaluate complete scripts from creative queries, we combine their narrative and cinematic dimensions with query-fulfillment criteria from writing evaluation \citep{wu2025writingbench}.
The resulting twelve metrics span query fulfillment, narrative coherence, cinematic executability, and dramatic effectiveness, with three per category.
Metrics and scoring rules remain fixed across evaluations.

Each final checkpoint generates one script per test query.
GPT-5.4 assigns each metric an integer from 0 to 2 using only the query, anonymized script, and fixed criteria (Appendix~\ref{app:evaluation}).
We normalize category and overall scores to $[0,100]$ before averaging within types and equally across types.

\subsection{Main Results}

\begin{table}[!ht]
    \centering
    \small
    \caption{\textbf{\ours{} improves over matched static baselines.}
    Scores are query averages within each video type.
    Avg.\ gives equal weight to the three types.
    RL results average four seeds.
    Parentheses show gains over the corresponding static baseline at the same update level.}
    \label{tab:main-results}
    \setlength{\tabcolsep}{4.5pt}
    \renewcommand{\arraystretch}{1.08}
    \resizebox{\linewidth}{!}{%
    \begin{tabular}{lrrrr}
        \toprule
        Model & 2D animation & 3D/stop-motion & Live action & Avg. \\
        \midrule
        \rowcolor{morandiBlue}
        \multicolumn{5}{l}{\emph{Large-scale reference models}} \\
        DeepSeek-V4-Pro \citep{deepseek2026v4}
        & 86.50 & 85.59 & 86.10 & 86.06 \\
        Gemini 3.1 Pro Preview \citep{google2026gemini}
        & 82.12 & 81.99 & 78.42 & 80.84 \\
        GPT-5 (2025-08-07) \citep{openai2025gpt5}
        & \underline{89.86} & \underline{89.39} & \underline{89.35} & \underline{89.53} \\
        Kimi K2.5 \citep{moonshot2026kimi}
        & 85.46 & 84.42 & 82.61 & 84.16 \\
        Qwen3.8-Max \citep{qwen2026qwen38}
        & \textbf{91.05} & \textbf{91.15} & \textbf{91.02} & \textbf{91.07} \\
        Claude Sonnet 4.6 \citep{anthropic2026sonnet}
        & 68.79 & 66.80 & 66.70 & 67.43 \\
        GLM-5.2 \citep{glm52026}
        & 72.31 & 72.55 & 71.96 & 72.27 \\
        \midrule
        \rowcolor{morandiSage}
        \multicolumn{5}{l}{\emph{Adapted evolving-rubric baselines}} \\
        MPO \citep{kim2025evaluative}
        & 72.66 & 73.87 & 68.31 & 71.61 \\
        RLER \citep{shao2025drtulu}
        & 71.36 & 73.27 & 66.47 & 70.37 \\
        \midrule
        \rowcolor{morandiRose}
        \multicolumn{5}{l}{\emph{Training baselines}} \\
        Base
        & 65.51 & 64.63 & 60.83 & 63.66 \\
        SFT
        & 46.29 & 44.10 & 41.54 & 43.98 \\
        Task-Static
        & 75.41 & 74.44 & 68.59 & 72.81 \\
        Sample-Static
        & 74.70 & 72.94 & 68.29 & 71.98 \\
        \midrule
        \rowcolor{morandiSand}
        \multicolumn{5}{l}{\emph{\ours{} variants}} \\
        \textbf{\ours{} (Sample-level, \textit{Negative})}
        & \textbf{78.25} & \textbf{76.84} & \textbf{72.31} & \textbf{75.80 ($+3.82$)} \\
        \quad with \textit{Positive} replacements
        & 76.37 & 75.59 & 69.91 & 73.96 ($+1.98$) \\
        \quad with \textit{Mixed} replacements
        & \underline{77.09} & \underline{75.67} & \underline{71.13} & \underline{74.63 ($+2.65$)} \\
        \ours{} (Task-level)
        & 76.27 & 75.56 & 69.84 & 73.89 ($+1.08$) \\
        \bottomrule
    \end{tabular}}
\end{table}

\paragraph{Static Rubric Rewards Improve Script Generation}
Task-Static and Sample-Static outperform Base on all video types, with mean gains of 9.15 and 8.32 points (Table~\ref{tab:main-results}).
Task-Static leads by 0.83 points on average, so query-specific criteria alone provide no advantage in this comparison.
SFT scores zero on requirement coverage for over 80\% of outputs in each type, while its rising validation loss suggests overfitting.
In this setting, explicit rubric rewards improve request fulfillment more effectively than learning from reference scripts alone.

\paragraph{Rubric Evolution Improves over Matched Static Baselines}

Every \ours{} variant improves over its matched static baseline on all video types, with mean gains up to 3.82 points at sample level and 1.08 at task level.
The strongest sample-level variant gains 3.55 to 4.02 points across types, showing that its average improvement is distributed across all three.
All variants also outperform the adapted MPO and RLER baselines, whose average scores fall below their static counterparts.
With task-level update frequency matched, the gains depend on how the rubric is revised.
To isolate the effect of a task-level revision, we continue training from a shared checkpoint using either the original or revised rubric.
The revised rubric yields higher endpoint scores across all types, supporting its role in guiding subsequent learning under otherwise matched training conditions (Appendix~\ref{app:task-level-fork}).

\paragraph{\textit{Negative} Replacement Performs Best}

\textit{Negative} consistently outperforms \textit{Mixed} and \textit{Positive} under the same trigger, budget, and scoring procedure, with mean margins of 1.17 and 1.84 points.
It improves all four quality categories over Sample-Static (Appendix~\ref{app:detailed-results}) and surpasses task-level evolution, reversing the static baseline ordering.
In these experiments, selective revisions targeting flaws improve both request fulfillment and broader script quality.
Under this evaluation protocol, the resulting 4B policy also outperforms GLM-5.2 and Claude Sonnet 4.6 across all three types.

\section{Analysis}
\label{sec:analysis}
We analyze script training to connect criterion-level discrimination with revision content and independently evaluated quality. Additional tasks then assess applicability beyond scripts.

\subsection{Restoring Criterion-Level Signals}

We rescore the same eight scripts in 3,336 \textit{Negative} events to test whether replacements distinguish previously tied outputs. We normalize each script's summed replacement scores by their combined maximum. Headroom is 1 minus the group mean. Both headroom and variance start at zero.

\begin{figure}[!htbp]
    \centering
    \begin{minipage}[b]{0.57\linewidth}
        \centering
        \includegraphics[width=\linewidth]{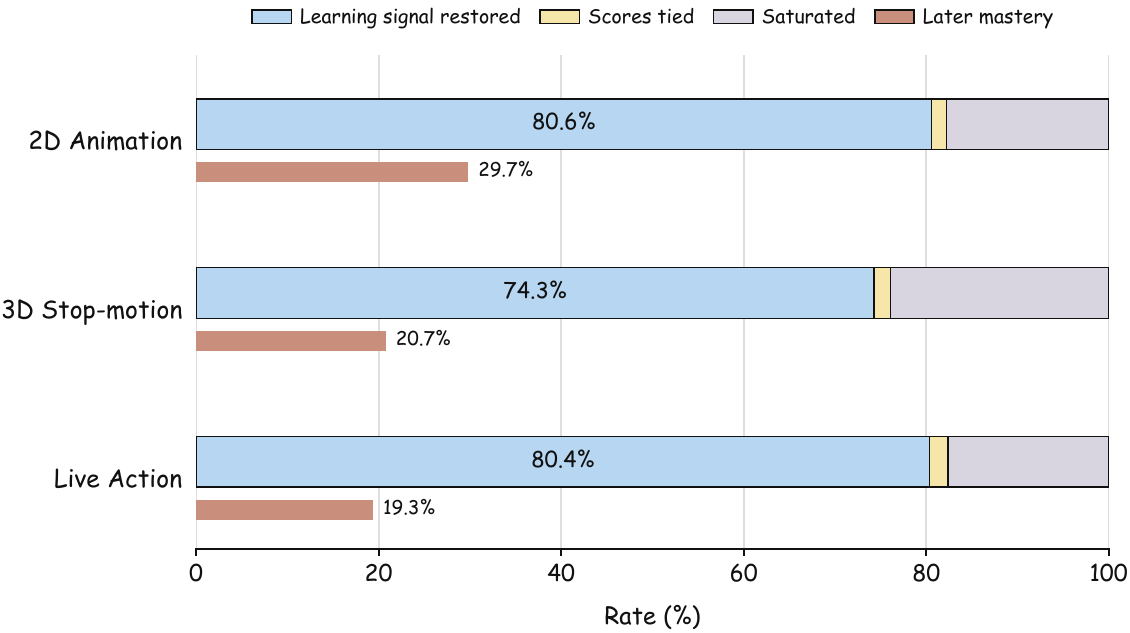}
        \par\smallskip\textbf{(a)}
    \end{minipage}\hfill
    \begin{minipage}[b]{0.41\linewidth}
        \centering
        \includegraphics[width=\linewidth]{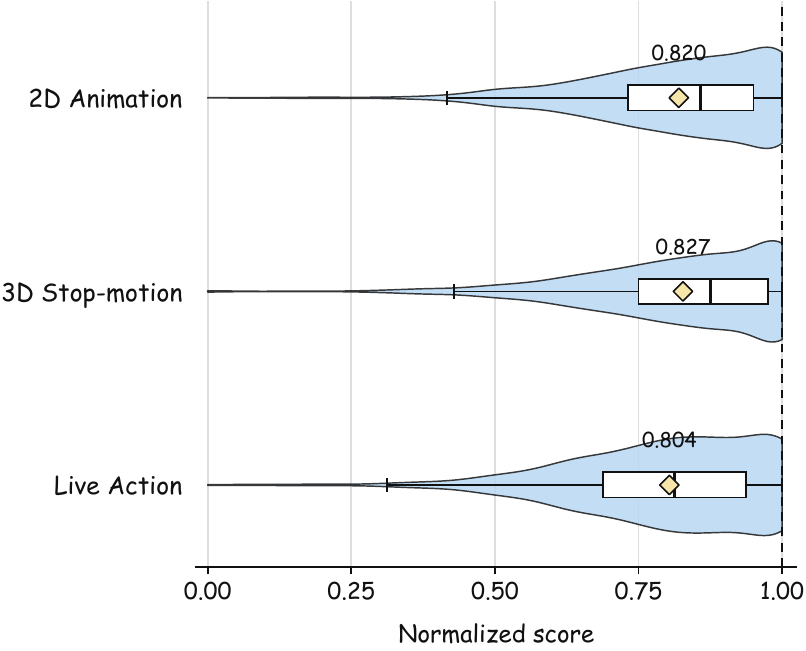}
        \par\smallskip\textbf{(b)}
    \end{minipage}
    \caption{\textbf{Replacements restore score differences.} Panel (a) shows separation and resaturation. In (b), boxes show interquartile ranges, inner lines medians, and diamonds means. Dashed lines mark full credit. Appendix~\ref{app:rubric-analysis} details the summaries.}
    \label{fig:criterion-signals}
\end{figure}

As shown in Figure~\ref{fig:criterion-signals}(a), 80.0\% of events reopen headroom and 78.2\% restore variance, pooling all video types. Uniform score reductions leave relative advantages unchanged. Panel (b) plots event means by video type, averaging normalized replacement scores over eight rollouts per event.

To examine later satisfaction, we track 846 criteria that reopen headroom and recur unchanged in epoch two. Of these, 207 award all eight scripts full credit, giving a pooled resaturation rate of 24.5\% across video types (Figure~\ref{fig:criterion-signals}(a)).
On newly sampled scripts from later checkpoints, new criteria reveal score differences more often than old criteria (Appendix~\ref{app:independent-replay}). \textbf{Replacements distinguish subsequent scripts while earlier requirements remain largely satisfied at both checkpoints.}

\subsection{Comparing Replacement Directions}

We compare replacement directions to assess whether targeting flaws provides more discriminative feedback. Across training runs, \textit{Negative} most often restores variance and yields the largest mean advantage magnitude (Table~\ref{tab:replacement-signals}(a)). These criteria also resaturate least often (Appendix~\ref{app:rubric-analysis}).

\begin{table}[!b]
    \centering
\caption{\textbf{\textit{Negative} replacements yield stronger signals.} In (a), $\hat A^{\mathrm{rep}}$ uses replacement scores only. In (b), $z$ standardizes each criterion separately. $\Delta$ gives relative changes, except percentage points (pp) for the headroom rate.}
    \label{tab:replacement-signals}
    \footnotesize
    \setlength{\tabcolsep}{2.2pt}
    \renewcommand{\arraystretch}{1.40}
    \begin{minipage}[t]{0.49\linewidth}
        \centering
        \renewcommand{\arraystretch}{1.20}
        \textbf{(a) Trigger events by direction}

        \begin{tabular}{lccc}
            \toprule
            Metric & \textit{Negative} & \textit{Mixed} & \textit{Positive} \\
            \midrule
            Post-switch score $\downarrow$ & \textbf{0.82} & 0.84 & 0.86 \\
            Opens headroom $\uparrow$ & \textbf{80.00\%} & 77.90\% & 68.50\% \\
            Restores variance $\uparrow$ & \textbf{78.20\%} & 76.40\% & 67.50\% \\
            Mean $|\hat A^{\mathrm{rep}}|$ $\uparrow$ & \textbf{0.66} & 0.65 & 0.57 \\
            Later resaturation $\downarrow$ & \textbf{24.50\%} & 36.30\% & 34.10\% \\
            \bottomrule
        \end{tabular}
    \end{minipage}\hfill
    \begin{minipage}[t]{0.495\linewidth}
        \centering
        \textbf{(b) Matched \textit{Mixed} events}

        \begin{tabular}{lccc}
            \toprule
            Metric & \textit{Positive} & \textit{Negative} & $\Delta$ \\
            \midrule
            Score headroom $\uparrow$ & 12.91\% & \textbf{15.80\%} & \textbf{+22.40\%} \\
            Variance $(\times 10^{-2})$ $\uparrow$ & 2.70 & \textbf{3.23} & \textbf{+19.40\%} \\
            Mean $|z|$ $\uparrow$ & 0.53 & \textbf{0.58} & \textbf{+8.40\%} \\
            Headroom rate $\uparrow$ & 64.48\% & \textbf{71.19\%} & \textbf{+6.71 pp} \\
            \bottomrule
        \end{tabular}
    \end{minipage}
\end{table}

We compare directions on identical scripts in 1,371 \textit{Mixed} updates. Criterion-level metrics are averaged within each direction, then across events (Appendix~\ref{app:rubric-analysis}). \textit{Negative} yields greater headroom and 19.4\% higher variance (Table~\ref{tab:replacement-signals}(b)).
\textbf{Flaw-focused criteria provide stronger discrimination.}

\subsection{Quality Requirements Targeted by Revisions}

\begin{figure}[!htbp]
    \centering
    \includegraphics[width=\linewidth]{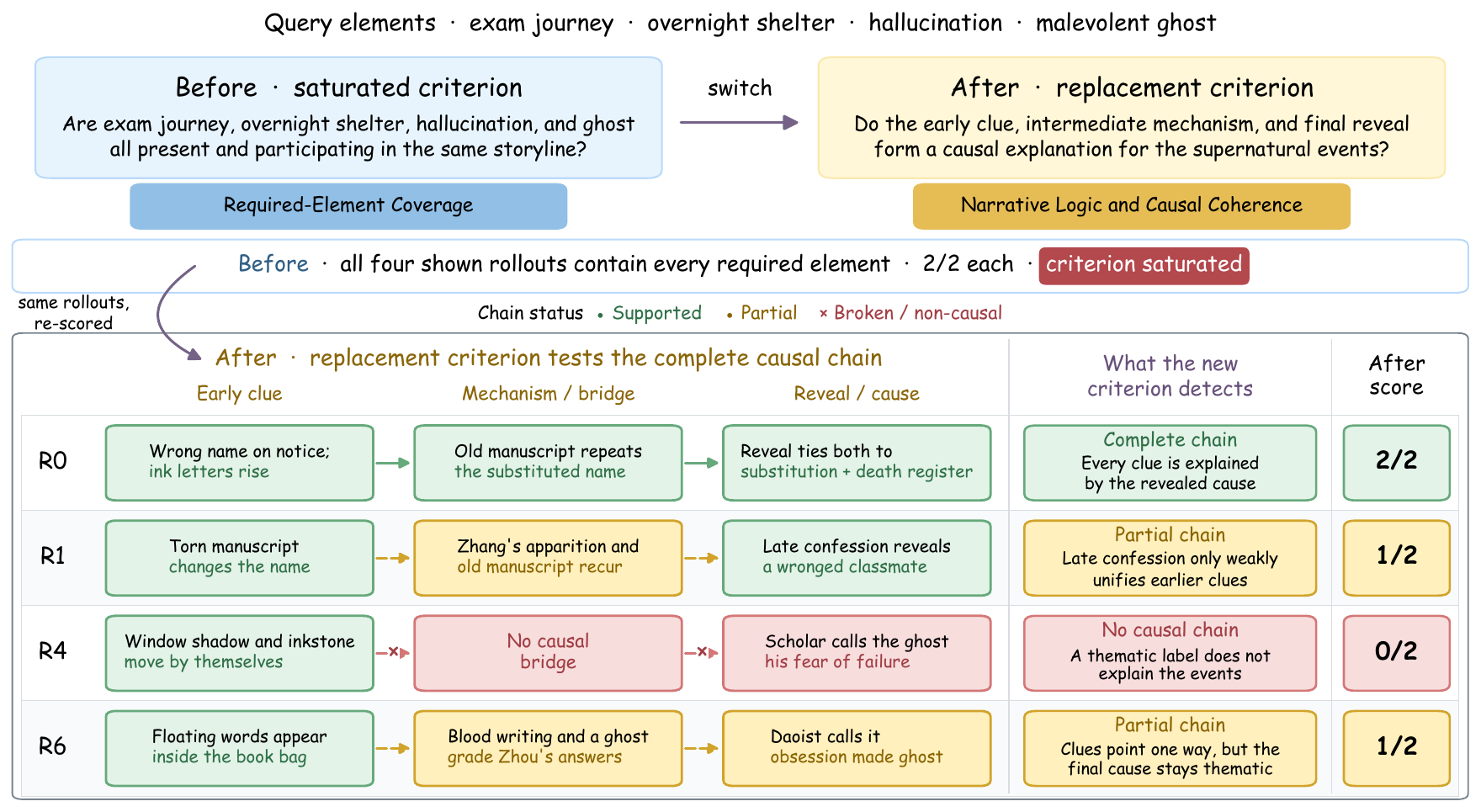}
    \caption{\textbf{A replacement exposes differences in causal coherence.} Four of eight unchanged scripts show how shifting from element coverage to causal explanation changes scores.}
    \label{fig:single-switch}
\end{figure}

We examine how revisions change criterion focus through the case study in Figure~\ref{fig:single-switch}. All eight scripts initially receive full credit for covering the query's required elements. Rescoring the same scripts under the replacement reveals differences in causal coherence. R0 retains full credit because its reveal explains preceding clues. R4 receives zero because naming the ghost a fear of failure leaves the supernatural events unexplained. R1 and R6 receive partial credit for incomplete explanations. \textbf{The replacement distinguishes narrative quality among scripts with identical coverage scores.}

\begin{figure}[!htb]
    \centering
    \begin{minipage}[b]{0.44\linewidth}
        \centering
        \includegraphics[width=\linewidth]{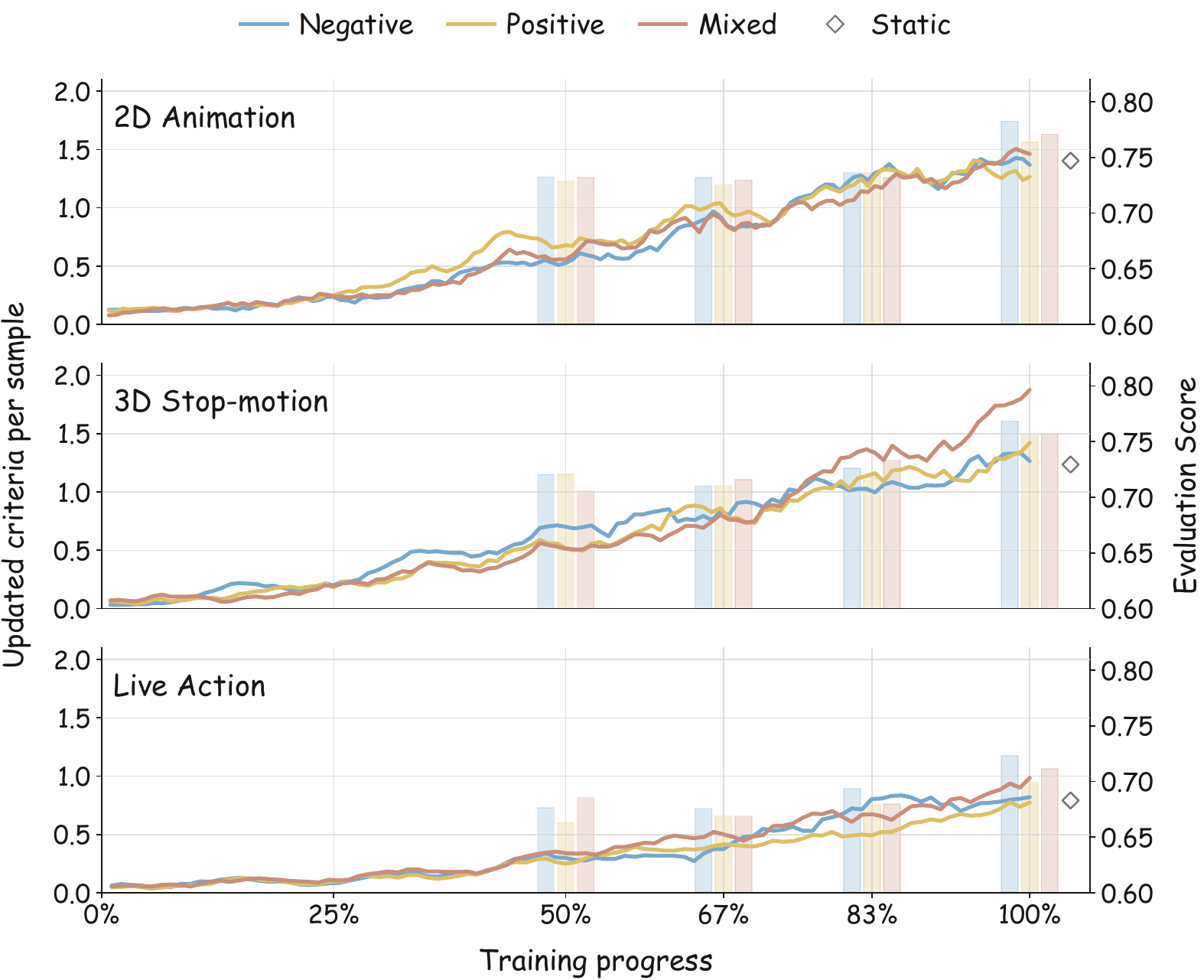}
        \par\smallskip\textbf{(a)}
    \end{minipage}\hfill
    \begin{minipage}[b]{0.54\linewidth}
        \centering
        \includegraphics[width=\linewidth]{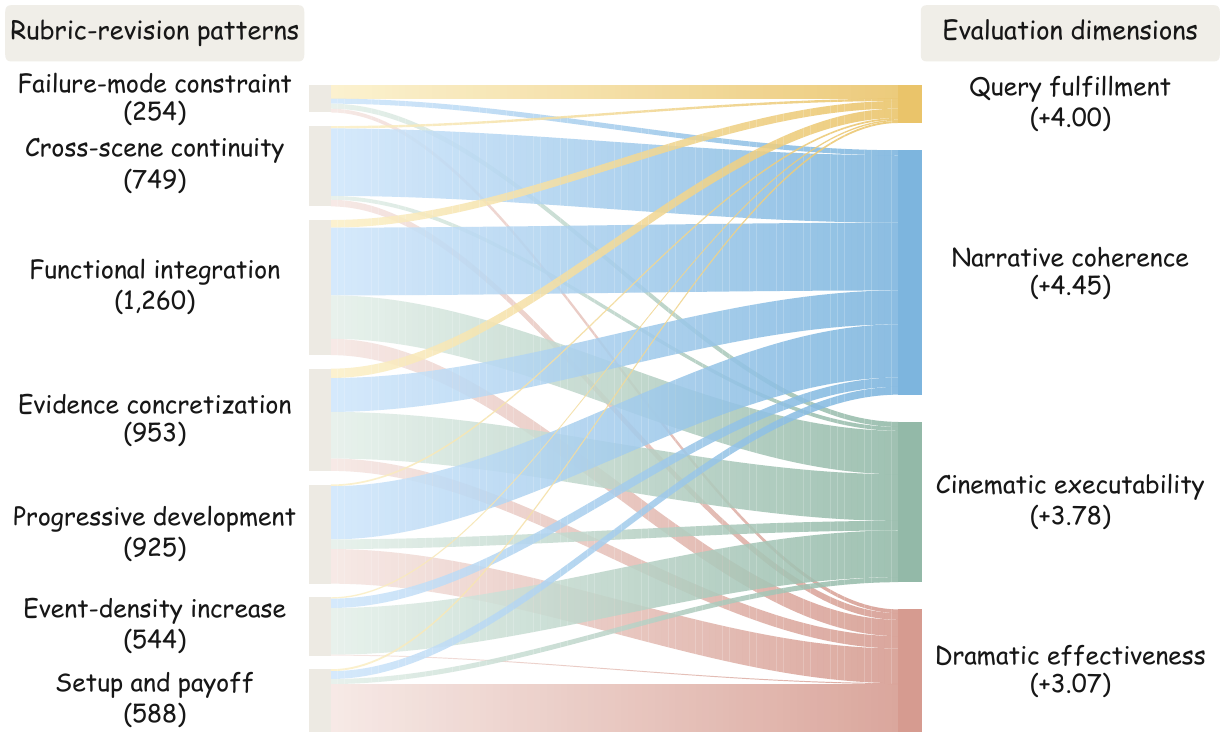}
        \par\smallskip\textbf{(b)}
    \end{minipage}
    \caption{\textbf{Rubric revisions accompany quality gains.} Panel (a) shows smoothed updates and evaluation scores. Panel (b) maps revision categories to evaluation dimensions.}
    \label{fig:revision-patterns}
\end{figure}

Across dynamic runs, more initial criteria reach full-score saturation as training progresses (Appendix~\ref{app:initial-saturation}). In Figure~\ref{fig:revision-patterns}(a), smoothed counts of replaced criteria per query (left axis) rise, consistent with the policy satisfying more current criteria. Fixed evaluation scores (right-axis bars, scaled to $[0,1]$) also improve. All variants finish above Sample-Static.

To characterize revision content, the judge classifies 5,273 \textit{Negative} replacements (Appendix~\ref{app:rubric-analysis}). Narrative function, concrete scene details, and progression account for 59.5\%. Figure~\ref{fig:revision-patterns}(b) links continuity to narrative coherence and setup and payoff to dramatic effectiveness. Shown gains average \textit{Negative}'s improvements over Sample-Static equally across video types. \textbf{Revisions target dimensions that also improve under fixed evaluation.}

\subsection{Pairwise Validation of Pointwise Evaluation}

\begin{figure}[!htb]
    \centering
    \includegraphics[width=0.92\linewidth]{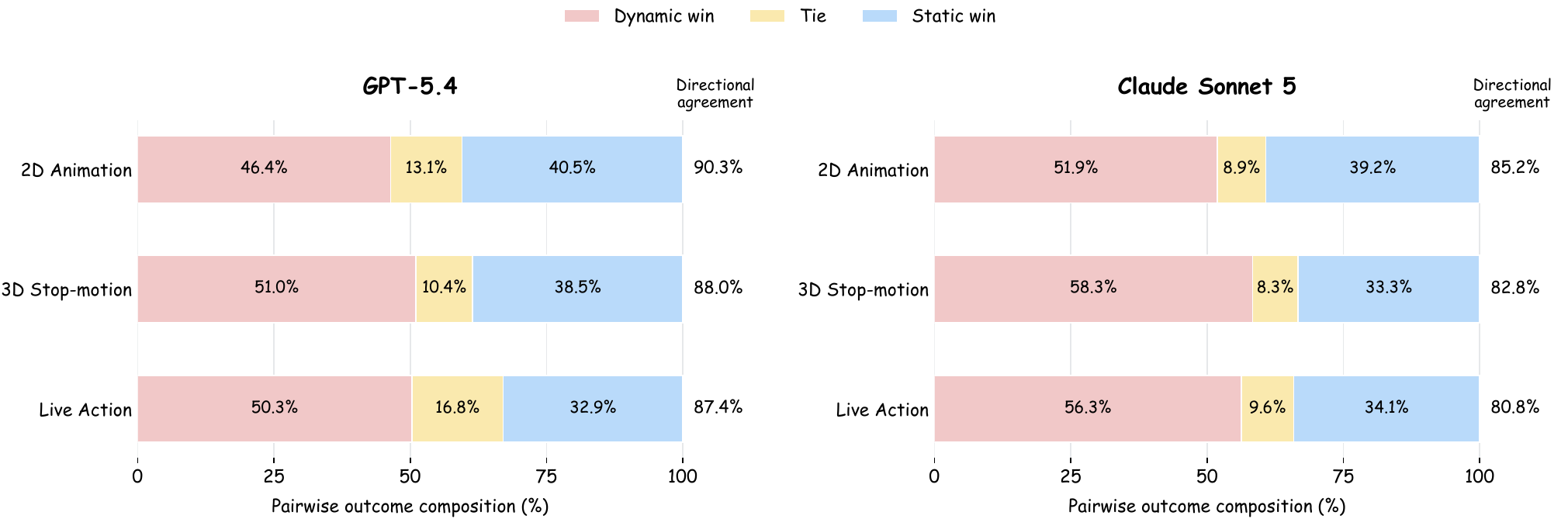}
    \caption{\textbf{Both judges favor \textit{Negative}.} Dynamic and Static denote \textit{Negative} and Sample-Static.}
    \label{fig:pairwise-preference}
\end{figure}

We test whether pairwise comparisons support the ranking from pointwise scores.
GPT-5.4 and Claude Sonnet 5 blindly compare \textit{Negative} and Sample-Static on 596 matched queries in both script orders (Appendix~\ref{app:evaluation}).
Figure~\ref{fig:pairwise-preference} shows more wins for \textit{Negative} under both judges in every video type. Over 80\% of preferences agree after reversing script order.

To assess human preference, five computer science PhD students blindly compare 60 script pairs, 20 per video type, with three independent ratings per pair (Appendix~\ref{app:human-evaluation}). Majority votes favor \textit{Negative} in 38 pairs and Sample-Static in 13, leaving nine ties or unresolved cases. Counting these as half a win yields a preference score of 70.8\% (95\% bootstrap confidence interval of 60.0\% to 80.8\%). \textbf{Human judgments support \textit{Negative}'s advantage in overall script quality.}

\subsection{Additional Text Generation Tasks}
\label{sec:generalization}
To assess applicability beyond scripts, we train separate policies on a WritingBench split for writing and Chasing Tail for general generation.
Each task compares sample-level \textit{Negative} with fixed per-query rubrics (Static) under matched data and training settings (Appendix~\ref{app:generalization}).
Both RL methods improve over Base under the official WritingBench \citep{wu2025writingbench} and BiGGen Bench \citep{kim2025biggen} protocols (Table~\ref{tab:external-generalization}).
\textbf{Rubric evolution outperforms static training on both tasks.}
These gains extend to tasks with different training distributions and evaluation criteria.

\begin{table}[!htbp]
    \centering
    \caption{\textbf{\ours{} outperforms Static on both tasks.} Scores are benchmark macro averages.}
    \label{tab:external-generalization}
    \small
    \renewcommand{\arraystretch}{1.15}
    \begin{tabular}{lcc}
        \toprule
        Model & BiGGen Bench (1 to 5) & WritingBench (1 to 10) \\
        \midrule
        Base & 4.035 & 8.22 \\
        Static & 4.152 & 8.26 \\
        \ours{} (\textit{Negative}) & \textbf{4.259} & \textbf{8.34} \\
        \bottomrule
    \end{tabular}
\end{table}

\section{Conclusion}
\label{sec:conclusion}
We introduced \ours{} to adapt reward feedback to uneven progress across requirements. Its sample-level procedure uses current scores to select criteria for revision and response differences to guide their replacements, retaining all other criteria and scores. Task-level revisions use representative outputs across queries. Script experiments show gains over matched static and adapted dynamic baselines, with flaw-focused replacements performing best. New criteria distinguish later scripts while earlier requirements remain largely satisfied. Improvements on two additional tasks support applying rubric evolution beyond scripts. Together, these findings link adaptive reward criteria with continued improvement in output quality.

\clearpage
\section*{AI Use Statement}

We used generative AI to aid and polish manuscript writing through translation, language editing, and formatting checks. For synthetic dataset generation, Gemini 3.1 Pro reconstructed scripts from source videos and GPT-5.4 generated the corresponding creative queries. We used Qwen3.7-Plus to score script training outputs and GPT-4.1-mini for the additional text tasks. GPT-5.4 generated and revised rubrics. It also scored the script benchmark pointwise and performed blind pairwise comparisons alongside Claude Sonnet 5. Evaluation models for WritingBench and BiGGen Bench are specified in Appendix~\ref{app:generalization}. The authors remain responsible for the manuscript and results.

\subsection*{Ethics statement}
This study uses source videos from the media collection adopted by AVA-Encoder and model-generated queries, rubrics, and scripts for model training and evaluation. Five computer science PhD students independently evaluate anonymized script pairs. All raters participated voluntarily and provided informed consent. Generated scripts may reflect biases or harmful stereotypes present in source media and pretrained models, so practical use requires human review.

\subsection*{Reproducibility statement}
The appendix reports data construction and source-video splits, model and judge versions, training and decoding settings, the evaluation protocol, detailed results, and rubric construction, scoring, and revision prompts. Main RL results average runs with seeds 0, 1, 2, and 42. Code is available \href{\paperCodeURL}{here}.

\bibliography{ref}
\bibliographystyle{iclr2027_conference}

\newpage
\appendix
\section{Implementation Details}
\label{app:implementation}

We describe the script experiments below, including their data, rubric initialization, and evaluation protocol. Configurations for additional text generation tasks appear in Appendix~\ref{app:generalization}.

\subsection{Script Data and Initial Rubrics}
\label{app:data-construction}

We construct query--script pairs from the media collection used by AVA-Encoder \citep{li2026ava}, covering 2D animation, 3D and stop-motion animation, and live action. Gemini 3.1 Pro \citep{google2026gemini} reconstructs a detailed script from each complete video. GPT-5.4 \citep{openai2026gpt54} then derives five to seven creative queries from that script. Each query expresses pre-production intent and omits the ending, character names, and timestamps.

To initialize rewards, GPT-5.4 compares each query's eight initial-policy scripts and constructs criteria from observable quality differences without prescribing candidate-specific plots. Each rubric contains $3\leq k\leq8$ criteria, including a fixed check of script content density. We summarize the query-specific rubrics into one shared six-criterion rubric for each video type.

We split the data by source video so that related queries and scripts remain together. All rubrics follow the split of their associated queries. Table~\ref{tab:dataset-statistics} reports the resulting training set of 5,393 queries from 837 videos and test set of 596 queries from 93 videos.

\subsection{Training and Rubric Evolution}
\label{app:training}

For SFT, we fine-tune Qwen3-4B-Instruct-2507 \citep{yang2025qwen3} separately for each production style. Each example contains a system instruction, query, and reconstructed target script. SFT runs for three epochs with a global batch size of 32 and a sequence limit of 8,192 tokens. We use an initial learning rate of $1\times10^{-5}$, a cosine schedule with 3\% warmup, and right truncation. Validation loss rises from the first to the third epoch, consistent with overfitting.

All script RL methods start from the initial Qwen3-4B-Instruct-2507 checkpoint and train for two epochs on each production style. GRPO uses a global batch size of 32 queries and samples $G=8$ scripts per query. The learning rate is $1\times10^{-6}$, with prompt and response lengths capped at 4,096 and 8,192 tokens. We average runs with seeds 0, 1, 2, and 42 for the main results.

The training judge, Qwen3.7-Plus, scores each criterion under the current rubric at temperature 0 with at most 4,096 output tokens. GRPO normalizes advantages by the within-group reward standard deviation. We optimize a token-mean dual-clipped PPO loss with clipping ratio 0.2 and dual-clip coefficient 3.0. The low-variance KL term has coefficient 0.001.

For each query occurrence, the adapted RLER baseline generates one to five criteria from the query, eight current rollouts, and the previously active adaptive pool. The judge scores the initial rubric, active pool, and new criteria together on these rollouts. The current reward sums all criterion scores and divides by their combined maximum, so new criteria contribute immediately.

After computing rewards, RLER removes adaptive criteria with zero population standard deviation across the eight rollouts. It retains up to five with the largest standard deviations for the query's next occurrence. Initial criteria are always scored and are excluded from filtering and the five-item pool limit. Inactive criteria are used only for auditing and deduplication.

\subsubsection{Sample-Level Rubric Updates}
\label{app:sample-updates}

Sample-level updates replace saturated criteria to restore score differences within a rollout group. Criterion $c_j^{(t)}$ is eligible when all $G$ rollouts have valid scores equal to its maximum $m_j$. At temperature 0.2, GPT-5.4 uses differences among these responses to generate up to $B=3$ replacements. The update preserves the number of criteria, their IDs, and their maximum scores, keeping their relative weights fixed. \textit{Positive}, \textit{Negative}, and \textit{Mixed} differ only in replacement direction.

The rubric generator returns exactly $b_t=\min(B,|\mathcal{S}_q^{(t)}|)$ replacements, each defining every integer score from $0$ to $m_j$. The prompt requires distinct evidence for new and retained criteria and excludes candidate-specific plot details. In \textit{Mixed}, one replacement can use either direction, while $b_t\geq2$ requires at least one criterion of each direction in the same update.

After rubric validation, the judge scores all replacement criteria on the same $G$ rollouts in one request. We validate the score matrix and save the revised rubric before applying all replacements together. New and retained scores form $\tilde r_i^{(t)}$ for the policy update. If no revision is triggered or retries fail, training retains the original rubric and uses $\bar r_i^{(t)}=r_i^{(t)}$.

For an accepted revision, the stored rubric changes according to
\begin{equation}
    c_j^{(t+1)}
    =
    \begin{cases}
        c_j^{\prime(t)}, & j\in\mathcal{U}_q^{(t)},\\
        c_j^{(t)}, & j\notin\mathcal{U}_q^{(t)}.
    \end{cases}
    \label{eq:criterion-transition}
\end{equation}
The revised version is used on subsequent visits to the query.

\subsubsection{Task-Level Rubric Updates}
\label{app:task-updates}

Task-Static and task-level \ours{} each use a shared six-criterion rubric per production style. To match update frequency, task-level \ours{} and MPO perform five evenly spaced revisions before training ends. Task-level \ours{} activates accepted revisions at the next step and preserves criterion IDs, order, maxima, and reward aggregation throughout these updates.

Each revision uses evidence from steps $\{t-2,t-1,t\}$. Eligible groups contain at least four valid rollouts, with one group retained per query. A revision requires at least 12 distinct queries. We select 12 to 16 groups, stratifying by mean-reward rank when the eligible set is larger. Within each group, the valid script closest to the median reward serves as its representative. The generator receives these scripts and their queries together with group means and criterion-level feedback.

We assess candidate rubrics for discrimination and changes in reward scale. After structural validation, the judge rescores the same representatives. We require valid paired scores for at least 12 representatives and at least two distinct new rewards. The difference between old and new mean normalized rewards must remain within $0.35$ in absolute value.

\subsubsection{API Call Overhead and Training Time}
\label{app:training-calls}

We quantify online API call overhead across one two-epoch GRPO run per production style. Counts cover successful training requests and exclude retries, data construction, initial rubric generation, validation, and final evaluation. Task-Static and Sample-Static each require 85,504 judge calls across the three styles. Task-level \ours{} adds 15 generator calls and 240 calls to rescore representatives, totaling 85,759 calls and a 0.30\% increase over Task-Static. Each sample-level \textit{Negative} event adds one generator call and one judge request that rescores all eight rollouts. Its 3,336 events bring the total to 92,176 calls, a 7.80\% increase over Sample-Static.

The training judge and rubric generator allow up to 12 and 16 concurrent API requests, respectively. Validation judging allows four concurrent requests. In the logged two-epoch runs, total active training time is 105.9 hours for 2D animation, 59.1 hours for 3D and stop-motion animation, and 40.4 hours for live action. Runtime varies mainly with output length and API latency.

\subsection{Evaluation Protocol}
\label{app:evaluation}

We evaluate each checkpoint by generating one script per test query at temperature 1.0 and top-$p$ 1.0, with at most 8,192 output tokens. The 596 queries follow the production-style split in Table~\ref{tab:dataset-statistics}.

GPT-5.4 scores the twelve fixed metrics in Table~\ref{tab:evaluation-metrics} using only the query and anonymized script. Evaluation uses temperature 0 and at most 32,768 output tokens. The judge first parses the query and checks script completeness. It then performs three fixed evidence checks and one adversarial check per metric before assigning $e_\ell\in\{0,1,2\}$ (Appendix~\ref{app:benchmark-prompt}). We aggregate three metrics per category, giving raw maxima of 6 per category and 24 overall. With $\mathcal{M}_c$ denoting the metrics in category $c$, normalization to a 100-point scale gives
\begin{equation}
    S_c=100\cdot\frac{\sum_{\ell\in\mathcal{M}_c}e_\ell}{3\times2},
    \qquad
    S_{\mathrm{overall}}=100\cdot\frac{\sum_{\ell=1}^{12}e_\ell}{12\times2}.
    \label{eq:evaluation-score}
\end{equation}

Before aggregation, an external validator checks the JSON schema, quoted evidence, and score arithmetic. Content scores use the twelve quality metrics, while formatting statistics are recorded separately. Within each seed, we average scores over queries in each production style and then equally across styles. Reported RL results average these scores across the four training seeds.

\begin{table}[!t]
    \centering
    \small
    \caption{\textbf{Twelve fixed metrics assess script quality.} Abbreviations match the detailed result tables.}
    \label{tab:evaluation-metrics}
    \renewcommand{\arraystretch}{1.15}
    \begin{tabularx}{\linewidth}{@{}>{\raggedright\arraybackslash}p{0.42\linewidth}X@{}}
        \toprule
        Metric & Assessment focus \\
        \midrule
        \rowcolor{morandiBlue}\multicolumn{2}{l}{\textbf{Query Fulfillment (QF)}} \\
        Requirement Coverage (Cov.) & Realizes the requested content and outcomes while satisfying explicit formal constraints. \\
        Grounded Elaboration (Elab.) & Develops the query through relevant additions without contradicting its requirements or diluting the main storyline. \\
        Tone and Style Alignment (Style) & Matches the requested genre, mood, rhythm, and visual style. \\
        \midrule
        \rowcolor{morandiSage}\multicolumn{2}{l}{\textbf{Narrative Coherence (NC)}} \\
        Causal and Temporal Logic (Logic) & Maintains plausible causal links and a consistent temporal sequence. \\
        Character Motivation and Behavior (Motiv.) & Aligns actions, dialogue, and character changes with goals, knowledge, and circumstances. \\
        Cross-Scene State Continuity (Cont.) & Maintains consistent states of characters, relationships, locations, props, and information across scenes. \\
        \midrule
        \rowcolor{morandiSand}\multicolumn{2}{l}{\textbf{Cinematic Executability (CE)}} \\
        Visual and Auditory Observability (Obs.) & Conveys key information through images, actions, dialogue, or sound. \\
        Shot Feasibility and Atomicity (Atom.) & Specifies clear, realizable shots without excessive or conflicting actions. \\
        Pacing and Shot Utility (Pace) & Uses appropriate shot duration, transitions, and information density, with a clear purpose for each shot. \\
        \midrule
        \rowcolor{morandiRose}\multicolumn{2}{l}{\textbf{Dramatic Effectiveness (DE)}} \\
        Engagement and Progression (Eng.) & Establishes interest at the opening and sustains development and change. \\
        Setup, Payoff, and Ending (Payoff) & Resolves earlier setups and completes or meaningfully reframes preceding events. \\
        Distinctiveness and Resonance (Dist.) & Avoids formulaic treatment and leaves a specific, story-relevant impression. \\
        \bottomrule
    \end{tabularx}
\end{table}

For pairwise validation, GPT-5.4 and Claude Sonnet 5 blindly compare \textit{Negative} and Sample-Static on 596 matched queries in both script orders. A third pass resolves conflicting judgments. Directional agreement is the fraction of judgments unchanged by script order (Figure~\ref{fig:pairwise-preference}).

\subsubsection{Human Evaluation}
\label{app:human-evaluation}

We assess human preferences using 60 pairs of \textit{Negative} and Sample-Static scripts. Within each test set, a fixed seeded hash ranking selects 20 distinct source videos and one query per video. Sampling is independent of model scores and comparison outcomes. We pair outputs for the same query from each method's final checkpoint trained with seed 1 for the corresponding video type.

Five computer science PhD students each evaluate 36 pairs, with 12 per video type and three independent ratings per pair. Model identities are hidden. For each rater and video type, \textit{Negative} appears first in six pairs and second in six, balancing model order within each assignment.

Raters read the query and both complete scripts, then choose the better script or a tie based on query fulfillment, story coherence, filmability, and dramatic and audiovisual effectiveness. They are instructed to use ties only when they cannot distinguish overall quality and to work independently. Only the overall preference is required, while written reasons and comments are optional.

At least two matching votes determine a pair's outcome. We assign one point to a \textit{Negative} win, zero to a Sample-Static win, and half a point to a tie or a case without a majority. Averaging over all 60 pairs gives the preference score. The 95\% confidence interval uses the 2.5th and 97.5th percentiles of 10,000 bootstrap replicates, resampling pairs with replacement within each video type. Nominal Krippendorff's $\alpha=0.71$ measures agreement among individual ratings.

\subsection{Rubric Analysis}
\label{app:rubric-analysis}

For Table~\ref{tab:replacement-signals}(a), we sum each rollout's replacement scores and divide by the combined maximum of those criteria. We standardize these normalized scores within each group of eight rollouts to obtain $\hat A_i^{\mathrm{rep}}$, then average their absolute values. This measures score separation among replacements. The policy update uses the full reward, which also includes retained criterion scores.

Figure~\ref{fig:criterion-signals}(b) summarizes event means by video type. Each event contributes one observation, averaging normalized replacement scores across eight rollouts. White boxes span the 25th to 75th percentiles, with an internal line at the median. Whiskers extend to the outermost observations within 1.5 interquartile ranges of the corresponding box edge. Outlier markers are omitted. Yellow diamonds and the values printed above them report the arithmetic mean across events. The dashed line at $1$ marks full credit, which every rollout received under the replaced criteria before revision.

To assess later satisfaction, we track criteria that reopen headroom in epoch one and recur unchanged in epoch two. The cohorts contain 846 \textit{Negative}, 815 \textit{Mixed}, and 783 \textit{Positive} criteria. Resaturation requires all eight scripts to receive full credit under the same unchanged criterion.

To compare directions on identical outputs, we separate the \textit{Positive} and \textit{Negative} criteria within each accepted \textit{Mixed} update. Of 3,223 updates from training seed 1 across the three video types, 1,371 contain both directions. Each pair shares the query, training step, generator call, judge, and eight rollouts. Across all 5,222 \textit{Mixed} replacements, 66.0\% are \textit{Positive} and 34.0\% are \textit{Negative}.

For each criterion, we normalize scores as $x_{i,j}=s_{i,j}/m_j$ and compute their mean $\mu_j$ and population standard deviation $\sigma_j$ over eight rollouts. Standardized scores are $z_{i,j}=(x_{i,j}-\mu_j)/(\sigma_j+10^{-6})$. The criterion contributes headroom $1-\mu_j$, variance $\sigma_j^2$, and mean absolute standardized score $G^{-1}\sum_i|z_{i,j}|$. For the headroom rate, each criterion contributes $1$ if $\mu_j<1-10^{-12}$ and $0$ otherwise.

Within each event, we average each statistic equally over criteria of the same direction. Table~\ref{tab:replacement-signals}(b) then averages events equally across the pooled video types. Relative changes use the two overall means, while the rate difference is reported in percentage points. Bootstrap estimates use 10,000 resamples of 603 source-video clusters defined within datasets, with random seed 20260831.

To characterize revision content, the training judge classifies 5,273 \textit{Negative} replacements using fixed revision categories and evaluation dimensions. Inputs contain the query, paired criteria, and generator's learning direction, excluding final evaluation scores. Narrative function, concrete scene details, and progression together account for 3,138 replacements (59.5\%). Figure~\ref{fig:revision-patterns}(b) aggregates these labels alongside gains over Sample-Static averaged equally across video types.

\subsubsection{Saturation of Initial Criteria During Dynamic Training}
\label{app:initial-saturation}

\begin{figure}[!htbp]
    \centering
    \includegraphics[width=\linewidth]{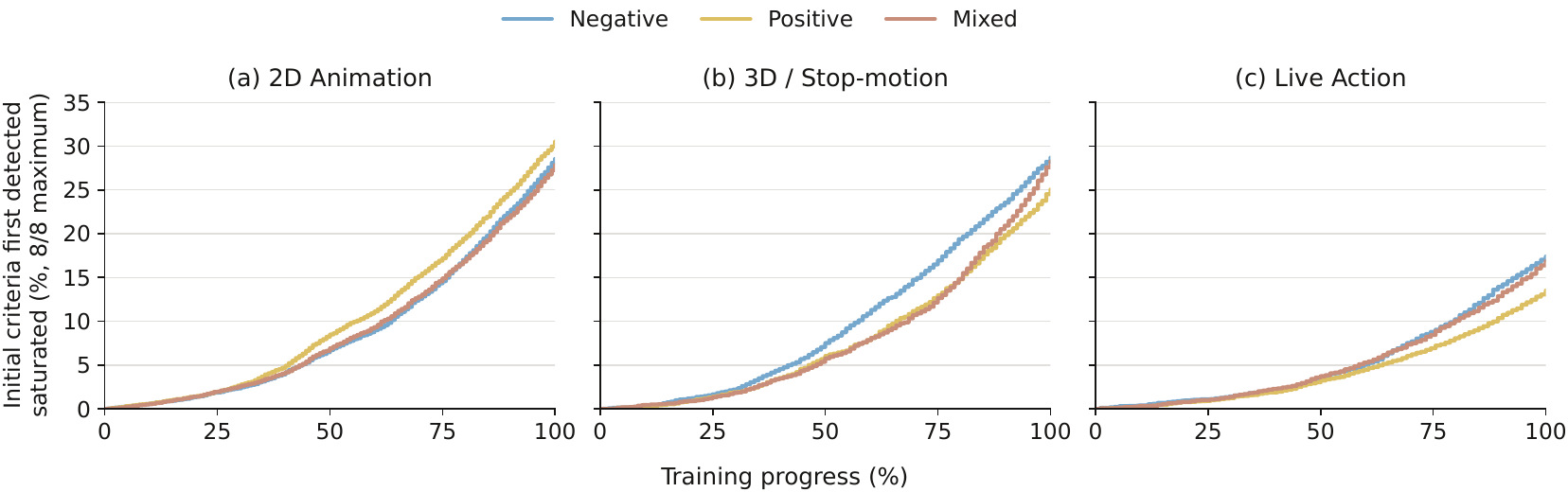}
    \caption{\textbf{Initial criteria reach saturation throughout dynamic training.} Curves show the cumulative fraction first awarding all eight rollouts full credit in each dynamic setting.}
    \label{fig:initial-saturation}
\end{figure}

We track initial criteria in the three dynamic settings to examine when they first lose discrimination. Figure~\ref{fig:initial-saturation} shows the cumulative fraction that has awarded all eight scripts full credit at least once. Each initial criterion enters the count at its first saturation, before replacement. The steeper rise later in training indicates that more initial criteria reach this state as the policy changes. These trajectories motivate checking criterion satisfaction throughout training.

\section{Local Effect of Task-Level Rubric Evolution}
\label{app:task-level-fork}

We compare training with the original and revised shared rubrics to isolate the effect of one task-level update. At the first accepted revision in each video type, both training branches start from checkpoint $\pi_{\theta_t}$ before the update. The revised rubric $\mathcal{R}^{(t+1)}$ uses evidence from the latest three steps of training with the original shared rubric.

Static retains the original rubric $\mathcal{R}^{(t)}$, while Updated activates $\mathcal{R}^{(t+1)}$ at the next step. Both rubrics remain fixed for 20 further steps. The paired continuations start at step 22 for 2D animation, 18 for 3D and stop-motion animation, and 16 for live action. Each pair shares the checkpoint, optimizer, scheduler, random state, and data order. GPT-5.4 independently evaluates the full test set before continuation and every five steps using the fixed metrics in Appendix~\ref{app:evaluation}.

\begin{figure}[!htbp]
    \centering
    \includegraphics[width=\linewidth]{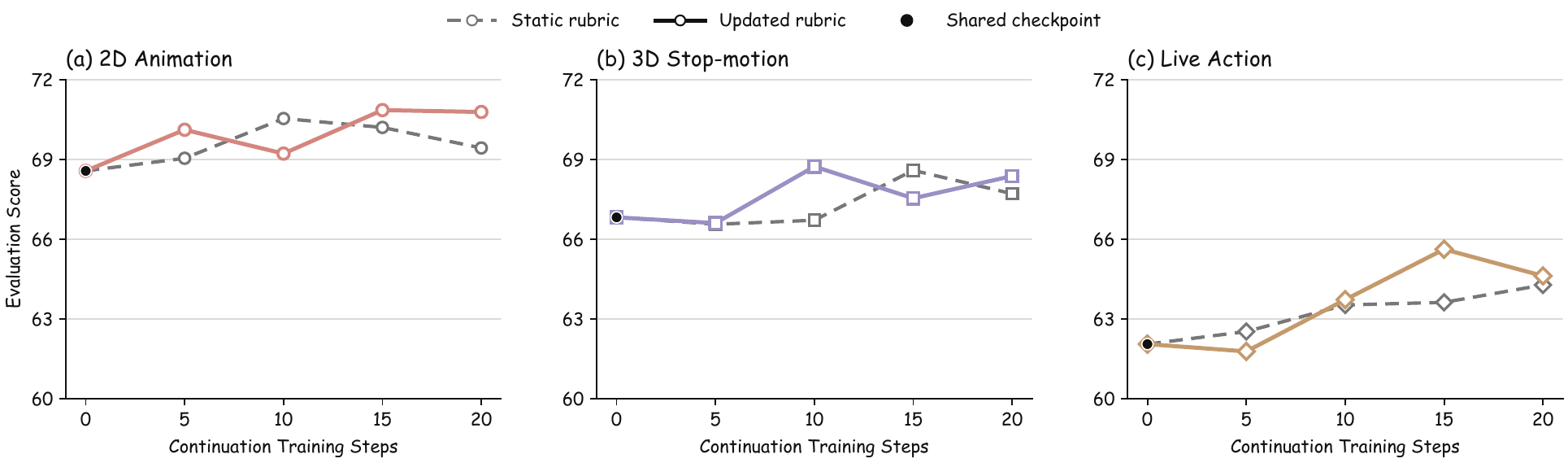}
    \caption{\textbf{A task-level revision improves subsequent training.} Curves compare branches trained for 20 steps from a shared checkpoint with fixed original or revised rubrics.}
    \label{fig:task-level-fork}
\end{figure}

Figure~\ref{fig:task-level-fork} shows higher final scores for Updated in all three production styles. Averaged equally across styles, its advantage over Static grows from 0.12 points at continuation step 5 to 0.79 points at step 20. Because the paired branches share their initial training state and data order, the comparison isolates the effect of the revised rubric. Updating the shared rubric therefore improves subsequent script quality at these three checkpoints under fixed evaluation metrics.

\section{Newly Sampled Scripts and Criterion Retention}
\label{app:independent-replay}

We test whether sample-level replacements distinguish newly generated scripts and whether these scripts still satisfy the replaced criteria. From the \textit{Negative} runs, we select accepted first-epoch replacements whose queries recur in epoch two. The resulting cohort contains 940 events from 939 unique queries, covering 1,210 pairs of old and new criteria. We aggregate all replacements within an event, using the same unit of analysis as the immediate comparison in Section~\ref{sec:analysis}.

For each query, we sample eight fresh scripts at the end of epoch one and again at the end of training. Each checkpoint therefore contributes 7,512 rollouts across production styles. Table~\ref{tab:independent-replay-domain} lists the corresponding training steps. Sampling uses temperature 1.0, top-$p$ 1.0, and at most 8,192 tokens, with task seeds matched between the epoch-boundary and final-checkpoint samples.

We compare old and new criteria on identical outputs. For each query and checkpoint, the training judge scores both versions jointly at temperature 0 with thinking disabled. For each rollout, we sum scores over the criteria replaced in an event and normalize by their combined maximum. Each event thus yields paired groups of eight normalized scores for comparing old and new criteria.

An event has headroom when its eight scores average below 1 and positive variance when their population variance exceeds zero. We assess retention per event and per rollout in Table~\ref{tab:independent-replay}. The column \textit{Old full groups} reports the fraction of events with full credit on every old criterion for all eight rollouts. \textit{Old full rollouts} measures the corresponding fraction of individual rollouts.

\begin{table}[!htbp]
    \centering
    \small
    \caption{\textbf{Revised criteria distinguish new scripts while earlier requirements remain largely satisfied.} Bold marks larger values across checkpoints.}
    \label{tab:independent-replay}
    \setlength{\tabcolsep}{4.2pt}
    \renewcommand{\arraystretch}{1.12}
    \resizebox{\linewidth}{!}{%
    \begin{tabular}{lrrrrrrr}
        \toprule
        Evaluation panel & Events & New mean & New headroom & New variance & Old full groups & Old mean & Old full rollouts \\
        \midrule
        Trigger batch (eligible cohort) & 940 & -- & 73.2\% & 72.0\% & 100.0\% & 1.000 & 100.0\% \\
        End of epoch 1 & 940 & 0.906 & \textbf{53.4\%} & \textbf{52.6\%} & 56.0\% & 0.943 & 86.6\% \\
        End of training & 940 & \textbf{0.931} & 43.4\% & 43.0\% & \textbf{66.3\%} & \textbf{0.957} & \textbf{90.4\%} \\
        \bottomrule
    \end{tabular}}
\end{table}

At both checkpoints, new criteria reveal headroom and score variation more often than paired old criteria, including events where old criteria remain saturated. Thus, replacements distinguish scripts beyond their original rollout group. From epoch one to the end of training, mean scores rise and full-credit groups become more common under the new criteria (Table~\ref{tab:independent-replay}). The policy increasingly satisfies these revised criteria on newly generated scripts.

\begin{table}[!htbp]
    \centering
    \scriptsize
    \caption{\textbf{Scores rise across all three production styles.} Metrics use newly sampled scripts at each checkpoint. Bold marks larger values within each style.}
    \label{tab:independent-replay-domain}
    \setlength{\tabcolsep}{3.0pt}
    \renewcommand{\arraystretch}{1.10}
    \resizebox{\linewidth}{!}{%
    \begin{tabular}{llrrrrrrr}
        \toprule
        Style & Checkpoint & Events & New mean & Headroom & Variance & Old full groups & Old mean & Old full rollouts \\
        \midrule
        2D animation & Epoch 1 (step 66) & 379 & 0.887 & \textbf{60.4\%} & \textbf{59.4\%} & 48.0\% & 0.928 & 83.5\% \\
        & Final (step 132) & 379 & \textbf{0.917} & 48.3\% & 47.8\% & \textbf{60.7\%} & \textbf{0.946} & \textbf{88.1\%} \\
        \midrule
        3D and stop-motion & Epoch 1 (step 54) & 382 & 0.926 & \textbf{46.6\%} & \textbf{46.1\%} & 63.4\% & 0.957 & 89.5\% \\
        & Final (step 108) & 382 & \textbf{0.944} & 40.1\% & 39.8\% & \textbf{70.2\%} & \textbf{0.968} & \textbf{92.3\%} \\
        \midrule
        Live action & Epoch 1 (step 47) & 179 & 0.906 & \textbf{53.1\%} & \textbf{52.0\%} & 57.0\% & 0.944 & 86.7\% \\
        & Final (step 94) & 179 & \textbf{0.931} & 40.2\% & 39.7\% & \textbf{69.8\%} & \textbf{0.958} & \textbf{90.9\%} \\
        \bottomrule
    \end{tabular}}
\end{table}

Earlier requirements remain largely satisfied as old-criterion scores and full-credit rates rise during training (Table~\ref{tab:independent-replay}). Table~\ref{tab:independent-replay-domain} shows the same pattern in each production style. The policy's improved scores under new criteria therefore accompany continued satisfaction of earlier requirements.

We repeat the analysis for the 1,210 criterion pairs to check the effect of aggregation. New criteria continue to expose score differences at both checkpoints, while mean old-criterion scores remain above 0.94. These results support the event-level findings on discrimination and retention.

\section{Switch-Budget Coverage and Group-Size Sensitivity}
\label{app:trigger-sensitivity}

We examine budget coverage and saturation detection in 3,336 \textit{Negative} events.

\paragraph{Switch budget.}
For a fixed response, only selected criteria change its reward, giving $|\tilde r_i^{(t)}-r_i^{(t)}|\leq(\sum_{j\in\mathcal{U}_q^{(t)}}m_j)/(\sum_{j=1}^{k}m_j)$. The bound depends on their combined weight. We use $B=3$. Before applying the cap, 196 events (5.88\%) contain more than three saturated criteria. In 94.12\% of events, all eligible criteria can therefore be replaced within the budget.

\paragraph{Group size.}
We randomly subsample the existing $G=8$ rescore matrices. False saturation occurs when all sampled scores reach the maximum despite a lower score in the full group. Its rate is 13.2\% for $G=4$ and 4.73\% for $G=6$. All eight rollouts recover the reference decision by construction. Smaller groups can therefore miss existing differences and trigger premature replacement. This analysis measures detection error on observed outputs.

\section{Additional Text Generation Tasks}
\label{app:generalization}

We train Qwen3-4B-Instruct-2507 separately on writing and general generation tasks. Each comparison includes Base, fixed per-query rubrics (Static), and sample-level \ours{} with \textit{Negative} replacements. Both RL methods share data, initial rubrics, and GRPO settings. Runs use training seed 1, a batch size of 32 queries, and $G=8$ rollouts. GPT-4.1-mini scores training outputs and replacements, while GPT-5.4 generates rubric revisions. Table~\ref{tab:external-generalization} reports results.

Task prompts refer to queries and candidate responses, assessing observable content, correctness, and instruction following. Sample-level evolution retains the same update rule. Saturation requires all eight judgments to be valid and equal to the criterion maximum. The generator replaces at most $B=3$ criteria and preserves their IDs and maxima. Rescoring the same outputs supplies new scores that combine with retained scores for the current policy update.

\subsection{WritingBench}

For open-ended writing, we use the WritingBench split released with RiT \citep{tian2026rit}, containing 500 training and 500 test prompts. Each training prompt supplies five initial binary criteria. Both RL methods train for 200 steps with a maximum prompt length of 24,567 tokens. Under the official WritingBench protocol \citep{wu2025writingbench}, Claude Sonnet 4.5 scores five criteria per response. Scores are averaged within responses, then within domains, and finally equally across domains.

\subsection{BiGGen Bench}

BiGGen Bench \citep{kim2025biggen} evaluates general generation. We train on the separate Chasing Tail generalist collection of 4,931 deduplicated queries with weighted binary criteria. Both RL methods train for two epochs, totaling 308 steps, with a maximum prompt length of 4,096 tokens.

For an initial criterion with weight $w_j$, dataset conversion sets $m_j=w_j$, and the judge prompt specifies scores of either $0$ or $w_j$. The initial reward is therefore $r_i=\sum_j w_j b_{i,j}/\sum_j w_j$, where $b_{i,j}$ indicates criterion satisfaction. A replacement retains $m_j$ but defines every integer score from $0$ through $w_j$, allowing ordinal feedback at the original weight scale. The score validator accepts integers throughout this range, while the initial binary scoring rule is specified in the judge prompt.

Prometheus 2 BGB 8$\times$7B (v2.0) scores responses from 1 to 5. The \href{https://github.com/prometheus-eval/prometheus-eval/blob/main/BiGGen-Bench/run_response_eval.py}{official Prometheus evaluation} excludes 18 instances from the absolute and relative LLM-judge tasks, leaving 747 of the benchmark's 765 instances. We average scores within nine capability groups and equally across groups.

\begin{sidewaystable}[p]
    \section{Detailed Results by Domain and Evaluation Dimension}
    \label{app:detailed-results}

    Tables~\ref{tab:2d-detailed}, \ref{tab:3d-detailed}, and~\ref{tab:live-detailed} report all twelve metrics for 2D animation, 3D and stop-motion animation, and live action. Table~\ref{tab:evaluation-metrics} defines the metrics and their abbreviations. \textit{Negative} improves every category mean over Sample-Static in each style. Bold and underlining mark the highest and second-highest values, with separate rankings for the reference models and Qwen3-4B policy variants in each metric column.

    \centering
    \footnotesize
    \setlength{\tabcolsep}{2.5pt}
    \caption{\textbf{\textit{Negative} improves all four quality categories over Sample-Static in 2D animation.} Entries report metrics and category means.}
    \label{tab:2d-detailed}
    \resizebox{\textheight}{!}{%
    \begin{tabular}{l*{17}{r}}
        \toprule
        & \multicolumn{4}{c}{QF} & \multicolumn{4}{c}{NC} & \multicolumn{4}{c}{CE} & \multicolumn{4}{c}{DE} & \\
        \cmidrule(lr){2-5}\cmidrule(lr){6-9}\cmidrule(lr){10-13}\cmidrule(lr){14-17}
        Model & Cov. & Elab. & Style & Avg. & Logic & Motiv. & Cont. & Avg. & Obs. & Atom. & Pace & Avg. & Eng. & Payoff & Dist. & Avg. & Overall \\
        \midrule
        \rowcolor{morandiBlue}\multicolumn{18}{l}{\emph{Large-scale reference models}} \\
        DeepSeek-V4-Pro & 96.20 & \underline{99.37} & 97.68 & 97.75 & 78.69 & 93.88 & 76.58 & 83.05 & 97.05 & 93.46 & 49.58 & 80.03 & 94.51 & 62.87 & 98.10 & 85.16 & 86.50  \\
        Gemini 3.1 Pro Preview & 80.59 & \textbf{99.79} & 89.24 & 89.87 & 82.70 & 94.94 & 80.38 & 86.01 & 97.89 & 86.71 & 40.93 & 75.18 & 88.19 & 53.38 & 90.72 & 77.43 & 82.12  \\
        GPT-5 (2025-08-07) & \underline{99.58} & \textbf{99.79} & \underline{99.37} & \underline{99.58} & 88.19 & \underline{98.10} & \underline{82.70} & \underline{89.66} & \textbf{99.16} & \textbf{95.99} & \underline{51.69} & \underline{82.28} & \textbf{99.16} & \underline{64.77} & \underline{99.79} & \underline{87.91} & \underline{89.86 } \\
        Kimi K2.5 & 98.95 & 98.31 & 97.89 & 98.38 & 76.37 & 92.41 & 75.11 & 81.30 & 95.57 & 85.86 & 50.00 & 77.14 & 95.36 & 60.13 & 99.58 & 85.02 & 85.46  \\
        Qwen3.8-Max & \textbf{99.79} & \textbf{99.79} & \textbf{99.58} & \textbf{99.72} & \underline{88.40} & \textbf{98.31} & 80.59 & 89.10 & \underline{98.52} & \underline{94.73} & \textbf{58.86} & \textbf{84.04} & \underline{98.31} & \textbf{75.74} & \textbf{100.00} & \textbf{91.35} & \textbf{91.05 } \\
        Claude Sonnet 4.6 & 36.92 & \textbf{99.79} & 67.93 & 68.21 & \textbf{90.72} & 96.20 & \textbf{88.61} & \textbf{91.84} & 97.89 & 67.09 & 18.57 & 61.18 & 67.09 & 26.16 & 68.57 & 53.94 & 68.79  \\
        GLM-5.2 & 66.24 & 98.95 & 78.06 & 81.08 & 63.08 & 81.43 & 75.53 & 73.35 & 84.39 & 76.16 & 35.44 & 65.33 & 78.48 & 47.47 & 82.49 & 69.48 & 72.31  \\
        \midrule
        \rowcolor{morandiSage}\multicolumn{18}{l}{\emph{Adapted evolving-rubric baselines}} \\
        MPO & 78.27 & 91.35 & 88.82 & 86.15 & \textbf{59.70} & 78.69 & \textbf{59.07} & \textbf{65.82} & \textbf{87.97} & 54.22 & 41.77 & 61.32 & 84.60 & 55.91 & 91.56 & 77.36 & 72.66  \\
        RLER & 83.76 & 88.40 & 91.35 & 87.84 & 54.64 & 77.85 & 48.95 & 60.48 & 81.65 & 55.49 & 44.09 & 60.41 & 87.76 & 49.58 & 92.83 & 76.72 & 71.36  \\
        \midrule
        \rowcolor{morandiRose}\multicolumn{18}{l}{\emph{Training baselines}} \\
        Base & 86.50 & 90.93 & 90.51 & 89.31 & 44.94 & 58.65 & 53.59 & 52.39 & 55.27 & 52.95 & 48.73 & 52.32 & 64.77 & 50.84 & 88.40 & 68.00 & 65.51  \\
        SFT & 13.92 & 17.93 & 55.06 & 28.97 & 34.18 & 48.31 & 47.89 & 43.46 & 52.74 & 53.80 & 42.19 & 49.58 & 63.08 & 48.73 & 77.64 & 63.15 & 46.29  \\
        Task-Static & 92.62 & 91.98 & 94.30 & 92.97 & 54.01 & 77.85 & \underline{55.49} & 62.45 & 82.07 & 57.17 & 50.00 & 63.08 & 90.08 & \textbf{59.92} & 99.37 & 83.12 & 75.41  \\
        Sample-Static & 97.47 & 91.77 & 96.84 & 95.36 & 52.32 & 78.48 & 47.89 & 59.56 & 78.06 & \underline{63.08} & 48.95 & 63.36 & 88.40 & 54.64 & 98.52 & 80.52 & 74.70  \\
        \midrule
        \rowcolor{morandiSand}\multicolumn{18}{l}{\emph{\ours{} variants}} \\
        \ours{} (Task-level) & 95.57 & 93.88 & 96.62 & 95.36 & 54.64 & \textbf{81.86} & 51.05 & 62.52 & \underline{83.97} & 56.96 & 50.84 & 63.92 & \underline{92.19} & 58.02 & \underline{99.58} & \underline{83.26} & 76.27  \\
        \textbf{\ours{} (Sample-level, \textit{Negative})} & \textbf{100.00} & \textbf{97.05} & \textbf{99.79} & \textbf{98.95} & \underline{58.23} & 78.69 & 54.85 & \underline{63.92} & 83.12 & \underline{63.08} & \textbf{52.53} & \textbf{66.24} & \textbf{93.25} & 58.65 & \textbf{99.79} & \textbf{83.90} & \textbf{78.25 } \\
        with \textit{Positive} replacements & \underline{97.89} & \underline{94.73} & 98.31 & 96.98 & 54.43 & 77.43 & 51.90 & 61.25 & 82.91 & \textbf{63.50} & 50.63 & \underline{65.68} & 91.14 & 54.22 & 99.37 & 81.58 & 76.37  \\
        with \textit{Mixed} replacements & \textbf{100.00} & 93.25 & \underline{98.95} & \underline{97.40} & 54.85 & \underline{80.17} & 50.63 & 61.88 & 83.33 & 60.76 & \underline{51.48} & 65.19 & \textbf{93.25} & \underline{58.86} & \underline{99.58} & \textbf{83.90} & \underline{77.09 } \\
        \bottomrule
    \end{tabular}}
\end{sidewaystable}

\begin{sidewaystable}[p]
    \centering
    \footnotesize
    \setlength{\tabcolsep}{2.5pt}
    \caption{\textbf{\textit{Negative} improves all four quality categories over Sample-Static in 3D and stop-motion animation.} Entries report metrics and category means.}
    \label{tab:3d-detailed}
    \resizebox{\textheight}{!}{%
    \begin{tabular}{l*{17}{r}}
        \toprule
        & \multicolumn{4}{c}{QF} & \multicolumn{4}{c}{NC} & \multicolumn{4}{c}{CE} & \multicolumn{4}{c}{DE} & \\
        \cmidrule(lr){2-5}\cmidrule(lr){6-9}\cmidrule(lr){10-13}\cmidrule(lr){14-17}
        Model & Cov. & Elab. & Style & Avg. & Logic & Motiv. & Cont. & Avg. & Obs. & Atom. & Pace & Avg. & Eng. & Payoff & Dist. & Avg. & Overall \\
        \midrule
        \rowcolor{morandiBlue}\multicolumn{18}{l}{\emph{Large-scale reference models}} \\
        DeepSeek-V4-Pro & 95.57 & 98.96 & 95.57 & 96.70 & 78.12 & 94.27 & 79.43 & 83.94 & 95.31 & 88.80 & 48.96 & 77.69 & 94.27 & 59.64 & 98.18 & 84.03 & 85.59  \\
        Gemini 3.1 Pro Preview & 81.51 & 99.48 & 90.10 & 90.36 & 79.17 & 95.57 & \underline{80.47} & 85.07 & 96.35 & 83.07 & 42.71 & 74.04 & 89.84 & 54.17 & 91.41 & 78.47 & 81.99  \\
        GPT-5 (2025-08-07) & \underline{99.48} & \textbf{100.00} & \underline{98.96} & \underline{99.48} & \underline{87.24} & \underline{97.40} & 79.43 & 88.02 & \underline{98.70} & \underline{94.53} & \underline{53.65} & \underline{82.29} & \underline{99.22} & \underline{64.32} & \underline{99.74} & \underline{87.76} & \underline{89.39 } \\
        Kimi K2.5 & 97.40 & 97.92 & 97.14 & 97.49 & 71.09 & 93.23 & 73.96 & 79.43 & 92.71 & 84.11 & 50.00 & 75.61 & 93.49 & 61.98 & \textbf{100.00} & 85.16 & 84.42  \\
        Qwen3.8-Max & \textbf{99.74} & 99.22 & \textbf{100.00} & \textbf{99.65} & 85.68 & \textbf{99.22} & 79.95 & \underline{88.28} & \textbf{99.22} & \textbf{96.09} & \textbf{59.38} & \textbf{84.90} & \textbf{99.48} & \textbf{75.78} & \textbf{100.00} & \textbf{91.75} & \textbf{91.15 } \\
        Claude Sonnet 4.6 & 33.33 & \underline{99.74} & 65.36 & 66.14 & \textbf{87.50} & 96.88 & \textbf{83.33} & \textbf{89.24} & 97.14 & 66.93 & 17.19 & 60.42 & 65.62 & 21.88 & 66.67 & 51.39 & 66.80  \\
        GLM-5.2 & 67.45 & \underline{99.74} & 78.91 & 82.03 & 60.16 & 83.59 & 73.70 & 72.48 & 83.59 & 76.04 & 37.76 & 65.80 & 77.34 & 48.70 & 83.59 & 69.88 & 72.55  \\
        \midrule
        \rowcolor{morandiSage}\multicolumn{18}{l}{\emph{Adapted evolving-rubric baselines}} \\
        MPO & 74.74 & 89.58 & 86.72 & 83.68 & \textbf{65.89} & \textbf{83.85} & \textbf{64.58} & \textbf{71.44} & \textbf{88.28} & 58.07 & 44.01 & 63.45 & 85.16 & 52.08 & 93.49 & 76.91 & 73.87  \\
        RLER & 92.19 & 89.32 & 91.67 & 91.06 & 53.12 & 77.60 & 50.78 & 60.50 & 80.21 & 54.69 & 48.70 & 61.20 & 89.84 & 53.65 & 97.40 & 80.30 & 73.27  \\
        \midrule
        \rowcolor{morandiRose}\multicolumn{18}{l}{\emph{Training baselines}} \\
        Base & 79.69 & 85.42 & 87.76 & 84.29 & 41.41 & 56.51 & 54.17 & 50.70 & 54.95 & \textbf{69.53} & 47.92 & 57.47 & 60.42 & 50.00 & 87.76 & 66.06 & 64.63  \\
        SFT & 11.72 & 16.41 & 43.49 & 23.87 & 34.64 & 47.14 & 48.18 & 43.32 & 52.86 & 50.26 & 42.19 & 48.44 & 61.46 & 48.18 & 72.66 & 60.77 & 44.10  \\
        Task-Static & 90.36 & 90.62 & 91.67 & 90.88 & 53.39 & 73.18 & 52.86 & 59.81 & 82.29 & \underline{65.36} & 49.74 & \underline{65.80} & 86.20 & \textbf{58.33} & \underline{99.22} & 81.25 & 74.44  \\
        Sample-Static & 93.75 & 88.02 & 91.67 & 91.15 & 51.82 & 74.22 & 50.52 & 58.85 & 77.86 & 59.38 & 49.74 & 62.33 & 87.24 & 53.65 & 97.40 & 79.43 & 72.94  \\
        \midrule
        \rowcolor{morandiSand}\multicolumn{18}{l}{\emph{\ours{} variants}} \\
        \ours{} (Task-level) & 91.93 & \textbf{93.23} & 93.75 & 92.97 & 54.43 & 76.30 & 53.39 & 61.37 & 80.47 & 64.84 & \textbf{52.60} & \textbf{65.97} & 90.36 & 56.77 & 98.70 & 81.94 & 75.56  \\
        \textbf{\ours{} (Sample-level, \textit{Negative})} & \textbf{97.66} & \underline{92.97} & \textbf{96.09} & \textbf{95.57} & \underline{58.07} & \underline{77.86} & 52.86 & \underline{62.93} & 82.55 & 62.50 & \underline{52.08} & 65.71 & \textbf{92.45} & \underline{57.81} & \underline{99.22} & \textbf{83.16} & \textbf{76.84 } \\
        with \textit{Positive} replacements & \underline{95.83} & 91.67 & 95.05 & \underline{94.18} & 54.95 & 77.08 & 52.86 & 61.63 & 80.21 & 60.68 & 51.56 & 64.15 & \underline{91.15} & 56.25 & \textbf{99.74} & \underline{82.38} & 75.59  \\
        with \textit{Mixed} replacements & 95.05 & 92.19 & \underline{95.31} & \underline{94.18} & 54.69 & 77.60 & \underline{54.43} & 62.24 & \underline{84.38} & 63.02 & 50.00 & \underline{65.80} & 87.24 & 55.73 & 98.44 & 80.47 & \underline{75.67 } \\
        \bottomrule
    \end{tabular}}
\end{sidewaystable}

\begin{sidewaystable}[p]
    \centering
    \footnotesize
    \setlength{\tabcolsep}{2.5pt}
    \caption{\textbf{\textit{Negative} improves all four quality categories over Sample-Static in live action.} Entries report metrics and category means.}
    \label{tab:live-detailed}
    \resizebox{\textheight}{!}{%
    \begin{tabular}{l*{17}{r}}
        \toprule
        & \multicolumn{4}{c}{QF} & \multicolumn{4}{c}{NC} & \multicolumn{4}{c}{CE} & \multicolumn{4}{c}{DE} & \\
        \cmidrule(lr){2-5}\cmidrule(lr){6-9}\cmidrule(lr){10-13}\cmidrule(lr){14-17}
        Model & Cov. & Elab. & Style & Avg. & Logic & Motiv. & Cont. & Avg. & Obs. & Atom. & Pace & Avg. & Eng. & Payoff & Dist. & Avg. & Overall \\
        \midrule
        \rowcolor{morandiBlue}\multicolumn{18}{l}{\emph{Large-scale reference models}} \\
        DeepSeek-V4-Pro & 98.80 & 99.40 & 96.71 & 98.30 & 71.86 & 96.11 & 74.25 & 80.74 & 95.21 & \textbf{94.01} & 50.30 & 79.84 & 96.71 & 60.18 & \underline{99.70} & 85.53 & 86.10  \\
        Gemini 3.1 Pro Preview & 75.15 & 99.10 & 86.23 & 86.83 & 74.85 & 93.11 & 75.45 & 81.14 & 96.71 & 79.94 & 38.02 & 71.56 & 85.93 & 48.80 & 87.72 & 74.15 & 78.42  \\
        GPT-5 (2025-08-07) & \underline{99.40} & \underline{99.70} & \underline{99.10} & \underline{99.40} & \textbf{87.43} & \underline{97.60} & 79.64 & \underline{88.22} & \textbf{99.10} & \underline{93.41} & \underline{51.20} & \underline{81.24} & \textbf{99.70} & \underline{66.47} & 99.40 & \underline{88.52} & \underline{89.35 } \\
        Kimi K2.5 & 97.90 & 94.31 & 96.71 & 96.31 & 67.66 & 89.52 & 66.47 & 74.55 & 90.42 & 87.72 & 50.30 & 76.15 & 91.32 & 58.98 & \textbf{100.00} & 83.43 & 82.61  \\
        Qwen3.8-Max & \textbf{99.70} & \textbf{100.00} & \textbf{99.70} & \textbf{99.80} & \underline{85.63} & \textbf{98.80} & \underline{81.44} & \textbf{88.62} & \underline{97.90} & \underline{93.41} & \textbf{58.68} & \textbf{83.33} & \underline{99.40} & \textbf{77.54} & \textbf{100.00} & \textbf{92.31} & \textbf{91.02 } \\
        Claude Sonnet 4.6 & 34.43 & 99.40 & 66.47 & 66.77 & 85.03 & 96.71 & \textbf{81.74} & 87.83 & 95.81 & 65.87 & 17.37 & 59.68 & 65.57 & 24.55 & 67.37 & 52.50 & 66.70  \\
        GLM-5.2 & 65.87 & \underline{99.70} & 77.25 & 80.94 & 63.17 & 82.63 & 71.86 & 72.55 & 85.93 & 77.25 & 35.03 & 66.07 & 77.84 & 44.61 & 82.34 & 68.26 & 71.96  \\
        \midrule
        \rowcolor{morandiSage}\multicolumn{18}{l}{\emph{Adapted evolving-rubric baselines}} \\
        MPO & 87.72 & 84.43 & 85.33 & 85.83 & 47.01 & 63.17 & \textbf{50.60} & 53.59 & 67.37 & 62.57 & 43.11 & 57.68 & 84.13 & \underline{53.29} & 91.02 & 76.15 & 68.31  \\
        RLER & 83.53 & 74.55 & 82.34 & 80.14 & 46.11 & 64.37 & 48.80 & 53.09 & 70.66 & 57.49 & 47.01 & 58.39 & 80.54 & 50.60 & 91.62 & 74.25 & 66.47  \\
        \midrule
        \rowcolor{morandiRose}\multicolumn{18}{l}{\emph{Training baselines}} \\
        Base & 78.44 & 78.74 & 79.34 & 78.84 & 33.53 & 52.69 & \underline{50.30} & 45.51 & 51.80 & \textbf{70.66} & 45.51 & 55.99 & 58.68 & 49.40 & 80.84 & 62.97 & 60.83  \\
        SFT & 9.58 & 11.08 & 44.91 & 21.86 & 26.95 & 44.61 & 48.20 & 39.92 & 49.70 & 46.11 & 39.82 & 45.21 & 58.98 & 47.31 & 71.26 & 59.18 & 41.54  \\
        Task-Static & 89.22 & 82.63 & 87.43 & 86.43 & 44.91 & 63.47 & 44.61 & 51.00 & 69.46 & 61.98 & 49.70 & 60.38 & 80.54 & 52.10 & \textbf{97.01} & 76.55 & 68.59  \\
        Sample-Static & \textbf{95.21} & 79.64 & 85.33 & 86.73 & 43.11 & 63.17 & 48.20 & 51.49 & 68.26 & 55.99 & 49.10 & 57.78 & \underline{85.63} & 51.50 & 94.31 & 77.15 & 68.29  \\
        \midrule
        \rowcolor{morandiSand}\multicolumn{18}{l}{\emph{\ours{} variants}} \\
        \ours{} (Task-level) & 92.22 & 84.13 & 85.33 & 87.23 & \underline{48.50} & 64.07 & 48.80 & 53.79 & 67.37 & 62.87 & \textbf{51.80} & 60.68 & \textbf{88.32} & 51.50 & 93.11 & 77.64 & 69.84  \\
        \textbf{\ours{} (Sample-level, \textit{Negative})} & 94.01 & \textbf{86.53} & \textbf{91.62} & \textbf{90.72} & \textbf{50.30} & \textbf{68.86} & 50.00 & \textbf{56.39} & \textbf{73.95} & 64.67 & 50.00 & \textbf{62.87} & 84.43 & \textbf{56.29} & \textbf{97.01} & \textbf{79.24} & \textbf{72.31 } \\
        with \textit{Positive} replacements & \underline{94.61} & \underline{84.73} & 86.53 & 88.62 & 47.60 & 65.27 & 45.51 & 52.79 & 66.77 & \underline{64.97} & 50.30 & 60.68 & 83.53 & \underline{53.29} & \underline{95.81} & 77.54 & 69.91  \\
        with \textit{Mixed} replacements & \textbf{95.21} & 83.53 & \underline{90.12} & \underline{89.62} & 48.20 & \underline{67.07} & 48.20 & \underline{54.49} & \underline{71.86} & 64.37 & \underline{50.90} & \underline{62.38} & 85.03 & \underline{53.29} & \underline{95.81} & \underline{78.04} & \underline{71.13 } \\
        \bottomrule
    \end{tabular}}
\end{sidewaystable}

\clearpage

\section{Qualitative Comparisons of Generated Scripts}
\label{app:script-comparisons}

We compare \textit{Base}, the policy before RL, with sample-level \ours{} using \textit{Negative} replacements.
The six pairs contain two examples per video type, selected for higher overall pointwise scores and no decrease in any quality category after training.
Scene numbers follow the originals, and headings identify excerpts.
Other scenes retain their full narration and dialogue.
Separate sound and camera notes are omitted.
Bold marks the requirements and evidence used in each comparison.

\begingroup
\setlength{\tabcolsep}{7pt}
\setlength{\LTcapwidth}{\linewidth}
\renewcommand{\arraystretch}{1}

\newenvironment{scriptcomparison}[4]{%
    \begin{longtable}{p{\dimexpr\linewidth-2\tabcolsep\relax}}
        \caption{#1}\label{#2}\\
        \toprule
        \rowcolor{morandiBlue}
        \textbf{User query}\quad #3 \\
        \midrule
        \textbf{Comparison}\quad #4 \\
        \midrule
        \endfirsthead
        \multicolumn{1}{l}{\small Table~\thetable\ (continued)}\\
        \toprule
        \endhead
        \midrule
        \multicolumn{1}{r}{\footnotesize Continued on the next page}\\
        \endfoot
        \bottomrule
        \endlastfoot
}{%
    \end{longtable}
}
\newcommand{\baseheading}{%
    \rowcolor{morandiRose}
    \textbf{\textit{Base}}\\*
}
\newcommand{\negativeheading}{%
    \rowcolor{morandiSage}
    \textbf{\ours{} (\textit{Negative})}\\*
}
\newcommand{\scriptscene}[1]{%
    \textnormal{#1}\newline
}

\subsection{2D Animation}

We examine how each premise shapes the action.
Table~\ref{tab:case-chicken} concerns the requested usefulness of the chicken's long legs.
Table~\ref{tab:case-tree} traces the connection between possession and harm.

\begin{scriptcomparison}
    {\textbf{The chicken's long legs help recover missing toys.} Selected scenes from a 2D fairy tale.}
    {tab:case-chicken}
    {Could you write an imaginative fairy tale in which a chicken with exceptionally long legs \textbf{proves very useful at night}? It should be cute, absurd, and a little moving.}
    {\textit{Base} presents the long legs through moonlit footsteps and a reflection. \textit{Negative} uses them to solve the children's problem. The chicken presses the toy box with its long leg, returning the toys and Xiaohe's kite. This action fulfills the requested usefulness at night.}
    \baseheading
    \scriptscene{Scene 2, the first appearance, night}
    \textit{``As it walks, its toes lift the moonlight, as though writing a little unsigned poem.''}\\
    Moonlight covers the grass. The long-legged chicken slowly emerges from behind an old locust tree.
    \textbf{Each step stretches its shadow across the ground, like wind moving over a canvas.}
    The ends of its legs glow faintly, as if touched by stars.\\
    Xiaohe peers out from indoors, her voice trembling. ``You\ldots{} you're not a chicken?''\\
    \textit{``Not a chicken, but a bearer of the night.''}\\
    \scriptscene{Scene 3, guarding in the rain, night}
    \textit{``Only when the rain stopped did I realize it had been there all along, like the light and shadow I had lost.''}
    Raindrops tap the roof tiles. A puddle reflects the long-legged chicken.
    \textbf{It stands beneath a low eave and raises one foot until its toe touches the reflection in the water.}\\
    Xiaohe is drawing indoors. Her pencil pauses over a picture of a long-legged chicken standing in the rain.
    Xiaohe whispers, ``Aren't you afraid I'll draw you wrong?''\\
    \textit{``No. What I fear is forgetting that you will turn back.''}\\
    \negativeheading
    \scriptscene{Scene 3, the night of lost things, night}
    \textit{``My kite went missing yesterday. I searched all night, thinking the wind had taken it, until I saw that chicken step into the corner as though it were flying a kite.''}\\
    In front of the village toy house, the children raise the alarm. Their kites, matchboxes, and colored beads have all vanished.
    The village head grips a megaphone and shouts, ``Night ghosts! Who dares touch the children's things?''
    People stay behind closed doors.\\
    Atui, the long-legged chicken, appears in a corner and slowly approaches the wall. It taps the ground with its toe, and a faint glow appears on the paving stones.
    \textbf{Without a word, it rests its long leg on the edge of the toy box and presses lightly with its toe. The lid springs open, the toys return to their places, and a kite traces an arc through the air back to where it belongs.}\\
    \textbf{Xiaohe clutches her broken kite and bursts into tears of surprise and joy.}
    For the first time, she believes that something is watching over them at night.
    She asks Atui, ``Do you know what I lost?''
    Atui watches her quietly, then lifts its toe and lightly touches the back of her hand, as gently as patting someone on the back.
    After putting their toys away, the children say, ``It didn't make a sound, but it seemed to be listening.''\\
\end{scriptcomparison}

\clearpage
\begin{scriptcomparison}
    {\textbf{Possessiveness leads to visible harm.} Selected scenes from a 2D nature fable.}
    {tab:case-tree}
    {I would like a beautiful but somewhat cruel nature fable about why people should not casually \textbf{claim things that do not belong to them}.}
    {\textit{Base} moves from a gentle encounter to a severed trunk, then has Lin Yuan acknowledge turning the tree into an exhibit. \textit{Negative} depicts the intervention itself. Lin Ye's fear of losing the tree motivates binding and hammering its roots, which shake as the soil collapses. The scene links possession to harm through the character's actions.}
    \baseheading
    \scriptscene{Scene 3, a pact beneath the canopy, afternoon}
    \textit{``When I reached out to touch it, it reached back to hold my hand.''}\\
    The camera advances from Lin Yuan's perspective. He stands in the forest as sunlight falls through the leaves, the canopy covering him like an umbrella.
    He slowly reaches out his right hand and touches the trunk with his fingertips.
    The bark's patterns are as distinct as growth rings. The tree's shadow moves toward him across the ground, as if responding.\\
    His breathing slows. Lin Yuan says calmly, ``So it does move. It was simply waiting for me, for someone willing to believe in it.''\\
    \scriptscene{Scene 4, a man-made fracture, night}
    \textit{``They say a saw can save a tree. But what the saw cuts open is its memory.''}\\
    \textbf{Under the laboratory's harsh white lights, Lin Yuan stands before the severed trunk.}
    The saw teeth gleam, and the heartwood lies cut open. Its inner channels resemble ruptured blood vessels, with yellow sap slowly seeping out.
    He reaches out to touch it. The heartwood pulses faintly, as though breathing.
    His voice trembling, almost a whisper, Lin Yuan says, ``It hurts\ldots{} as much as when my research proposal was rejected for the first time.''\\
    \scriptscene{Scene 5, the tree falls silent, late at night}
    \textit{``I heard it say that you took away my name, you took away the way I breathe.''}\\
    Lin Yuan crouches beside the roots. The trunk has split, exposing its decaying interior.
    Moonlight passes through the heartwood, which has become transparent as glass.
    A faint light flows inside it like liquid starlight.
    He hears wind passing through the split, a slow, echoing whisper.
    His voice breaking, almost lost, Lin Yuan says, \textbf{``I\ldots{} only wanted to understand it\ldots{} but I turned it into an exhibit.''}\\
    \negativeheading
    \scriptscene{Scene 2, carving and harvesting, midday (excerpt)}
    \textit{\textbf{``I have to prove I own it.''}}\\
    \textbf{Lin Ye uses an iron-handled knife to carve ``Third Generation of the Lin Family'' into the south side of the trunk.}
    As the blade passes, the bark gives a faint crack, like a joint opening.
    He picks the first fruit, its skin amber-colored and containing points of light.\\
    \scriptscene{Scene 3, a cage around the roots, night}
    \textit{\textbf{``It will run away. I have to lock it down.''}}\\
    Late at night, Lin Ye brings a shovel and steel mesh and descends three meters beneath the tree roots.
    \textbf{He wraps wire mesh around three main roots and secures it with wooden stakes.}
    Iron nails enter the bark with a creak.
    \textbf{He strikes the roots with a hammer, saying, ``You cannot move. You belong to me.''}
    \textbf{The roots shake violently, the soil collapses, and deep cracks open.}
    The ground rises like a wound.
    The wind stops. Lin Ye shouts, ``Did you hear me? You cannot move anymore!''
    The roots jerk three times in succession. A crack an inch long spreads upward from underground, oozing thick black liquid into the soil.\\
    He lights a torch to inspect the roots and discovers a cluster of tiny glowing insects crawling over them.
    He swats at them in panic. The insects scatter, and black liquid slides out like blood.
    He swallows a piece of soil and mutters, ``The old rule says people must not claim the mountain. But this tree is a dead thing. If I take it all away, it should die.''
    He reinforces the wire mesh and chains, convinced that it can never be taken from him again.\\
\end{scriptcomparison}

\clearpage
\subsection{3D and Stop-Motion Animation}

We examine companionship under danger.
Table~\ref{tab:case-companion} contrasts dialogue and warning signals with physical shelter, while Table~\ref{tab:case-rescue} traces concern becoming action during a hazardous crossing.

\begin{scriptcomparison}
    {\textbf{Protective movement conveys companionship.} A request for dialogue-free animation.}
    {tab:case-companion}
    {Write a fairy-tale animation \textbf{without dialogue} about a vulnerable creature and something that cannot speak \textbf{keeping each other company in a dangerous environment}.}
    {\textit{Base} uses speech and later a glowing warning signal. In \textit{Negative}, the companion reshapes its body to shelter the beetle from a storm. Physical protection gives companionship a concrete function in the dangerous setting without character dialogue.}
    \baseheading
    \scriptscene{Scene 1, meeting on the snowfield, dusk}
    \textit{``I had never seen snow that could move. I thought it was merely a silent stone.''}\\
    Xiaoxue comes to a stop in the wind, its hind legs slightly bent and its ears twitching.
    Snow drifts from its nose and falls at its feet like fallen flowers.
    It looks up and sees a mound of snow shaped like a pillar broken by the wind.
    Moss covers its surface, and it sways slightly, as though waiting for Xiaoxue to approach.\\
    \textbf{Xiaoxue, softly, almost inaudibly, says, ``Can you\ldots{} hear the wind too?''}\\
    \textit{``The wind slips out of the rock crevice like a letter. It says that something is waiting for you.''}\\
    \scriptscene{Scene 5, before the avalanche, dusk}
    \textit{``The wind is growing stronger. It is moving. It is getting brighter.''}\\
    The wind intensifies, and the snowfield shakes. Xiaoxue retreats deeper into the crevice.
    The snow mound begins to glow like a lamp lit at night.
    \textbf{The ice patterns accelerate into a swirl, as though sending a signal. Xiaoxue knows this is a warning.}\\
    It wants to flee, but its feet seem pinned down.
    It wants to approach the mound, whose surface trembles with each step as though responding.\\
    \negativeheading
    \scriptscene{Scene 3, a shared ordeal}
    A blizzard strikes suddenly. Within ten seconds, the wind triples in speed, and ice crystals form dense bands in the air, covering the rock face.
    Xiaohupo trembles in a crevice, its antennae fully withdrawn and its body pressed against the rock.
    With each breath, its icy shell gives a faint creak.
    The silent body begins to change shape. Its surface repeatedly cracks and closes, producing a low hum like an old clock sounding in a room.
    At 16 Hz, the sound is inaudible but can be sensed as vibration by the crystalline structure inside Xiaohupo.\\
    \textbf{It moves toward Xiaohupo and extends its body section by section into a half-arch, enclosing the beetle within its gaps.}
    Xiaohupo curls up in fear, but the silent body does not press down on it.
    \textbf{It tilts slightly behind the beetle, opening a tiny, irregular pocket of shelter from the wind.}
    Xiaohupo breathes within it, sensing warmth slowly passing through the body, like glass gradually warming in sunlight rather than a flame.\\
    The wind intensifies. An icicle breaks off above, a sign of the approaching avalanche.
    It hangs frozen in midair like a meteor before falling.
    It strikes the silent body with a cracking sound, shattering its surface into several pieces.
    Blue light bursts out and scatters in the air as fleeting, star-shaped points.
    Xiaohupo recoils in terror and hits the rock wall. Its icy shell cracks with a sharp snap.
    The silent body holds together and begins to reassemble from its fragments.
    One piece gleams like a mirror, reflecting Xiaohupo's formerly curled-up outline.
    It gathers the pieces and repairs the cracks, moving slowly, like an organism healing itself.
    Each joining produces a faint, echoing hum.
    Xiaohupo remains still within it. The amber markings on its body flash in time with the cracks, briefly forming a shared pattern as the two structures begin to resonate.\\
\end{scriptcomparison}

\clearpage
\begin{scriptcomparison}
    {\textbf{Concern motivates a dangerous crossing.} A first-person animal story.}
    {tab:case-rescue}
    {Create a small-animal story from a first-person perspective. It should begin with foraging and survival, then show a brief spark of affection making the animal \textbf{willing to take risks} or even sacrifice itself.}
    {Both scripts express concern for a vulnerable animal. In \textit{Base}, the storm prevents Graytail from acting on it. \textit{Negative} connects an earlier moment of attachment to repeated attempts to reach Darktail despite unstable ice. Frostfur's return after retreating and its eventual crossing give the requested risk-taking a visible progression.}
    \baseheading
    \scriptscene{Scene 5, fleeing the sandstorm, noon}
    \textit{``The wind is fire, a blade, the world's anger.''}\\
    A violent wind rises. A sandstorm closes in from every direction, blurring earth and sky.
    Graytail scrambles about, its claws leaving deep marks in the sand.
    It looks back several times and sees the rabbit's burrow covered in sand. The rabbit does not move.
    \textbf{It wants to rush back but cannot take a step.}
    The wind has blown it about like a dead leaf.\\
    \textbf{``I can't\ldots{} I can't let it disappear,'' I say, hoarsely and haltingly.}\\
    \textit{``If I die, it will never have my shadow.''}\\
    \negativeheading
    \scriptscene{Scene 3, meeting Darktail, dusk (excerpt)}
    Frostfur wants to leave, then stops. It remembers an old fox bitten by wolves last winter.
    Before dying, the fox looked at the sky and slowly closed its eyes without howling.
    Frostfur wonders whether she, too, is waiting for something.\\
    \textbf{It draws back a step, then advances half a step and touches her ankle with its nose.}
    Darktail shudders, and their eyes meet.
    Frostfur sees a silver ring of snowlight around her pupils, resembling the old scar on its own left paw.
    Its chest tightens, its throat constricts, and it freezes.\\
    \scriptscene{Scene 4, crossing the ice ditch, night}
    \textit{``The ice ditch is not a path I should cross. My right shoulder hurts, my left paw is split, and it will break. But looking into her eyes, I suddenly feel that perhaps I should not turn back.''}\\
    It is late at night. Frostfur stops at the edge of a campfire.
    Darktail is still where she was, curling into a smaller shape, her breathing shallow and her right leg twitching.
    It wants to return, but something weighs on its chest.
    It remembers running across three snow pits for a mouthful of meat and never acting for anyone else.
    This time, it wants to save her.\\
    It begins climbing toward the ice ditch. The fissure is three meters wide, and the slope is unstable.
    Frostfur scrapes the ice with its hind paws. Each step lands on a loose layer of frost.
    An old scar marks its left forepaw, which twitches with each step as the bone makes a faint sound.
    It clenches its teeth and lengthens its stride.
    On the third step, its front paw slips into the fissure, and the ice breaks with a crack.
    It jerks its head, pulls free, and retreats two steps, almost falling before leaning against the rock wall.
    It pants heavily, as if fleeing for its life.\\
    \textbf{But it does not leave. It moves forward slowly, braces the ice with its body, presses its hind legs against the crack, and tests its footing.}
    The ice makes a faint cracking sound. Its heart jumps, and it dares not advance.\\
    It turns and sees Darktail curled up in the snow again, her right leg extended as if reaching for the ground.
    \textbf{Frostfur clenches its teeth, bites through a dead branch, and throws it toward the ditch as a bridge.}
    It presses its body onto the branch and pushes with its forepaws, inching into the fissure.
    \textbf{The ice breaks and it slips, its breathing broken into three gasps, but it finally makes it across.}
    It looks back and remains motionless.\\
    Darktail opens her eyes and looks at it, her gaze meeting its earlier stare.
    She stays still, lifting her left paw slightly as if to say thank you.
    Frostfur slowly walks back and spits out the blood accumulated in its throat, offering it to her instead of water.\\
    She takes half a mouthful and immediately withdraws.\\
    It had intended to leave at once, but stops partway, looks at her, and whispers.
    Its voice hoarse and barely audible, Frostfur says, ``You're cold. I'll carry you.''\\
    She moves, pressing her body against her right leg, and gives a faint gasp, as if answering.\\
\end{scriptcomparison}

\clearpage
\subsection{Live Action}

Table~\ref{tab:case-monster} examines the requested fight through attacks and defensive responses.
Table~\ref{tab:case-romance} follows a quiet romance through exchanges of drawings and poetry.
Both comparisons ask whether character actions support the intended conflict or relationship.

\begin{scriptcomparison}
    {\textbf{The protagonist fights back against the monster.} A nighttime pursuit.}
    {tab:case-monster}
    {Give me the opening of a monster short film in a stark, realistic style, preferably starting with a nighttime escape and a \textbf{fight for survival}.}
    {\textit{Base} shows an escape followed by a threatening appearance in the mirror. \textit{Negative} turns the encounter into combat. Smoke briefly forces the monster back, and a later pipe strike makes it stagger before the wound closes. These reactions connect the protagonist's attempts to survive with consequences for his opponent.}
    \baseheading
    \scriptscene{Scene 1, fleeing through an alley, night}
    \textit{``I have never seen it, but I know it is watching me. It does not move. It simply waits.''}\\
    Li Yuan slides through the narrow passage, his backpack hitting the iron wall with a dull thud.
    He lowers his head, sweeping his flashlight across the ground and its patterns of broken glass.
    He hears his heartbeat in time with breathing from a distant doorway.
    He keeps moving. An old poster on the wall to his right splits, revealing wet darkness behind it.\\
    Li Yuan whispers, almost inaudibly, ``It isn't a living thing\ldots{} it's the weather.''\\
    \textit{``I run. It follows.''}\\
    \scriptscene{Scene 3, the abandoned building next door, night}
    \textit{``It is in the mirror. It did not come through the door. It has existed since the moment I saw it.''}\\
    Li Yuan climbs onto the sixth-floor balcony. Wind rushes in, making the iron railing groan.
    He turns and sees a shadow in the balcony mirror, its head facing him.
    It is a shape assembled from fragments rather than a person.
    \textbf{He steps back and shines his flashlight at the mirror. The reflected shadow suddenly reaches out, its bony knuckles breaking through the glass.}\\
    His voice trembling, Li Yuan says, ``It isn't moving\ldots{} I'm seeing it.''\\
    \textit{``We are not escaping. We are being seen.''}\\
    \negativeheading
    \scriptscene{Scene 2, the blocked path, night (excerpt)}
    Chen Mo's breath catches. \textbf{He sprays the vehicle with the smoke canister.}
    The gas spreads into a gray-white cloud. The green liquid briefly recedes.
    \textbf{The monster's head twitches three times, then withdraws into shadow.}
    He stands, panting and gripping the canister until his knuckles turn white.
    He stares blankly, as though deciding whether to flee.\\
    \scriptscene{Scene 3, the first fight, night}
    \textit{``It crawls instead of walking. Every step seems to consume the ground.''}\\
    Chen Mo rushes toward the metal guardrail, his arm brushing a chain, and slips.
    He rolls over, gets up, and accelerates toward the railing.
    \textbf{Suddenly, the shadow behind him tears across the ground like a fluid, extends two bristling arms, and lunges straight at him.}
    Chen Mo turns sharply, sweeps his arm across, and dives toward the automatic control panel.\\
    He breaks a wire, sending out sparks.
    During the attack, the monster's right arm strikes the guardrail with a sharp metallic crack.
    Shallow serrations appear on the surface, releasing thick gray-white liquid.
    The drops burst as they land, scattering gray powder.
    Chen Mo steps back, striking his left ankle against a steel bar. Blood pours out.
    Trembling, he pulls out the smoke canister and sprays it at the monster's head.
    The gas blackens its scalp and makes it twitch briefly, but mucus jets from both shoulder hollows as if liquid valves have opened.\\
    \textbf{Chen Mo kicks the control panel away and swings an iron pipe into its shoulder.}
    A metallic clang erupts.
    \textbf{The monster staggers backward, and a deep horizontal cut opens in its skin, seeping blood before immediately closing like rubber reforming.}\\
    Chen Mo pants, his pupils dilating.
    In his reflection on the guardrail, he sees an identical cut on his own right shoulder, glowing faintly gray-white.\\
\end{scriptcomparison}

\clearpage
\begin{scriptcomparison}
    {\textbf{The relationship develops through drawings and poetry.} A quiet romance.}
    {tab:case-romance}
    {Could you write a first-person romantic micro-script with a \textbf{quiet protagonist who likes drawing and writing}? The story should feel cinematic.}
    {\textit{Base} moves from a brief glance to the narrator's claim of being understood, with little response from the other character to support it. \textit{Negative} shows Su Rui responding to a drawing and later to Lin Yu's poem. Their relationship develops through these exchanges.}
    \baseheading
    \scriptscene{Scene 3, the empty space by the window, day}
    \textit{``You sit there longer than the sunset lasts.''}\\
    I return to the classroom and sit in my original seat.
    Sunlight falls on the windowsill. Chen Mo is looking down, gently touching his pencil to the draft paper.
    The lines are thin, as though tracing a breath.
    Looking down at the paper, I ask very softly, ``Are you drawing the wind? Or those moments when you look up each time I hold my pen?''\\
    I look up and see him raise his eyes, his gaze passing over me.
    \textbf{Without a word, he sets his pencil on the corner of the desk and turns away.}
    \textit{``I suddenly feel you are waiting rather than drawing. Waiting for me to put down my pen and pick it up again.''}\\
    \scriptscene{Scene 4, the forgotten desk corner, night}
    \textit{``I looked through seven sketchbooks and never found a second drawing. There was only a water-damaged page, with the figure's hand reaching toward me.''}\\
    Late at night, I sit alone in the classroom corner under dim yellow light.
    I open the old sketchbook. Water stains spread across it.
    On page five, the figure in the drawing reaches a hand toward me.
    My voice trembling, barely speaking, I say, \textbf{``Only then did I realize you were the first person to understand the loneliness in my drawings.''}\\
    \textit{``You think I am cold, but I have simply turned all my words into lines.''}\\
    \negativeheading
    \scriptscene{Scene 3, arranging the class exhibition, day (excerpt)}
    Su Rui approaches and gently takes her boat drawing down from the wall.
    \textbf{She places it beneath Lin Yu's painting and writes, ``You draw more than I dare to say.'' Without looking up, she puts the sheet in his sketchbook.}\\
    Lin Yu sees the words. His fingers tighten slightly, but he stays still.
    He opens his sketchbook and finds a previously covered sketch of Su Rui in a blue dress, walking down the corridor.
    It shows her from behind, at exactly the same angle as in the painting.\\
    \scriptscene{Scene 5, evening study, night}
    \textit{``I am afraid to say it, afraid she will suddenly leave, afraid the world will fall silent with only my written words remaining.''}
    Heavy rain falls outside.\\
    The classroom lights are dim and yellow, and the air is damp.
    Lin Yu opens an old notebook at the corner of his desk and writes a short poem.\\
    ``The wind passes through the corridor, and you do not turn back.\\
    Yet I see your shadow, like a flower without a name.''\\
    \textbf{He folds the paper into a little boat and puts it in Su Rui's sketchbook, beneath the page where she drew a boat two days earlier.}
    He thinks no one will see it.\\
    \textbf{But the next morning, Su Rui takes out the page, circles a short passage at the end of the poem in red, and writes, ``Your writing is gentler than I imagined.''}\\
    She never mentions it again, but that day she puts another drawing on the wall, a boat sailing toward dusk with the word ``First'' written on its stern.\\
    Lin Yu looks at the word. His fingers rest motionless on the edge of his sketchbook.\\
\end{scriptcomparison}

\endgroup

\clearpage

\section{Representative Prompts}
\label{app:prompts}

These condensed prompts cover rubric construction, training rewards, revision, and benchmark evaluation. Sample-level revision compares current rollouts, while task-level revision uses representative scripts across queries. Benchmark evaluation uses twelve fixed metrics. The complete benchmark prompt and output schema are stored in \path{prompts/script_benchmark_judge_v9.md}.

\refstepcounter{subsection}
\begin{promptbox}{\thesubsection. Initial Rubric Generation (abridged)}
\begin{lstlisting}
You are an expert screenplay evaluator. Construct a query-specific rubric
for scoring eight candidate scripts sampled from the initial policy.

Read the query, all candidate scripts, and the general screenplay-quality
dimensions. Identify dimensions with observable differences across the
scripts. Exclude dimensions that nearly all scripts satisfy to the same
degree or fail without any gradable difference.

Generate 3 to 8 independent criteria in total. Always include the fixed
criterion for content density in the screenplay body. Each criterion must:
1. assess one observable property of the screenplay text.
2. define every integer score level using distinct, quotable evidence.
3. separate the current scripts near the policy's capability boundary.
4. generalize beyond the current scripts without prescribing a plot.
5. avoid names, events, dialogue, or other details from the candidates.
6. use evidence different from every other criterion.

For a graded criterion, the highest score must require narrative function,
causal closure, progression, scene-level realization, or another observable
form of execution. Keyword mentions and metadata alone cannot earn the
highest score.

Return a short rationale followed by one JSON block:
{
  "query": "<original query>",
  "rubric_version": 1,
  "generated_at_step": null,
  "generated_from_trigger": "init",
  "criteria": [{
    "id": "<criterion id>",
    "check": "<one observable check>",
    "scoring_rule": "<all integer score levels>",
    "max_points": <int>
  }]
}

Inputs:
Query: {QUERY}
Candidate scripts: {ROLLOUTS_BLOCK}
General quality dimensions: {QUALITY_RULES}
\end{lstlisting}
\end{promptbox}

\refstepcounter{subsection}
\begin{promptbox}{\thesubsection. Training-Time Criterion Scoring (abridged)}
\begin{lstlisting}
Score the candidate script using the given rubric. Judge each criterion
independently and use only evidence that can be located in the script. Read
the check and scoring rule before selecting an integer score. For each
criterion, quote one short excerpt that supports the decision. Do not infer
the author's intent. When evidence falls between two levels, select the lower
level.

Return JSON only:
{
  "criteria_scores": [
    {"id": "<criterion id>", "score": <int>, "max": <int>,
     "reason": "<quoted evidence and rationale>"}
  ],
  "total_score": <sum of scores>,
  "total_max": <sum of maxima>
}

Inputs:
Query: {QUERY}
Rubric: {RUBRIC}
Candidate script: {RESPONSE}
\end{lstlisting}
\end{promptbox}

\refstepcounter{subsection}
\label{app:sample-prompt}
The three sample-level conditions share the prompt below and differ only in the text inserted into \texttt{\{POLARITY\_INSTRUCTION\}} (Table~\ref{tab:polarity-prompts}). The evidence format, output schema, and replacement constraints are identical for \textit{Negative}, \textit{Positive}, and \textit{Mixed}.

\begin{table}[!htbp]
    \centering
    \small
    \caption{\textbf{Polarity instructions for the shared prompt.} Translated from Chinese.}
    \label{tab:polarity-prompts}
    \renewcommand{\arraystretch}{1.15}
    \begin{tabularx}{\linewidth}{l>{\raggedright\arraybackslash}X}
        \toprule
        Condition & Instruction \\
        \midrule
        \textit{Negative} &
        Every update must use negative polarity. A response receives the maximum score when the target defect is absent. The score decreases progressively when the defect is partially present, and becomes zero when the defect is pronounced and harms the response quality. \\
        \midrule
        \textit{Positive} &
        Every update must use positive polarity. A response receives the maximum score when the target quality is fully realized. The score decreases progressively as the realization becomes weaker, and becomes zero when the quality is absent from the response. \\
        \midrule
        \textit{Mixed} &
        Each update may use either positive or negative polarity. If two or more updates are generated, the resulting set must contain both polarities. For either polarity, a higher score must always indicate higher response quality. \\
        \bottomrule
    \end{tabularx}
\end{table}

\begin{promptbox}{\thesubsection. Sample-Level Criterion Replacement (abridged)}
\begin{lstlisting}
Several criteria in the current rubric received full scores on every rollout
for the same query. Compare all rollouts. Identify {UPDATE_COUNT} clear and
independent quality differences, then map each difference to the most suitable
saturated criterion.

{POLARITY_INSTRUCTION}

Each replacement must preserve the selected criterion's id and max_points.
It must express one observable learning direction and define every integer
score from the maximum to zero. It must generalize beyond the current
rollouts and must not include rollout IDs, candidate-specific plot details,
or a prescribed realization. Different replacements and retained criteria
must use different evidence for increasing or decreasing a score.

Return JSON only:
{
  "updates": [{
    "id": "<saturated criterion id>",
    "polarity": "<positive or negative>",
    "diagnosis": {
      "difference": "<observable contrast>",
      "best_rollout_ids": ["<id>"],
      "weaker_rollout_ids": ["<id>"],
      "learning_direction": "<general direction>",
      "nonredundancy": "<difference from other criteria>"
    },
    "rubric": {
      "id": "<same criterion id>",
      "check": "<one observable check>",
      "max_points": <unchanged>,
      "scoring_rule": "<all integer score levels>"
    }
  }]
}

Inputs: {QUERY}, {TRIGGER_CONTEXT}, {CURRENT_RUBRIC},
{SATURATED_CRITERIA}, {RETAINED_CRITERIA},
{CURRENT_ROLLOUTS_WITH_SCORES}
\end{lstlisting}
\end{promptbox}

\refstepcounter{subsection}
\begin{promptbox}{\thesubsection. Task-Level Rubric Evolution (abridged)}
\begin{lstlisting}
Rewrite a dataset-level rubric using representative rollouts from 12 to 16
distinct queries collected over the latest three training steps. Read each
query, rollout, criterion score, and scoring rationale. Summarize the current
capabilities as mastered, unreachable_now, and current_frontier. Move the
scoring boundary toward observable capabilities that distinguish weaker,
adequate, and stronger scripts across queries.

Preserve the number and order of criteria, every id and max_points value, and
the total score. Each criterion must assess one independently improvable
capability and define every integer score level with evidence from the
screenplay body. Do not copy names, settings, events, dialogue, endings, or a
specific plot solution from the evidence cases.

Return JSON only:
{
  "task_id": "<unchanged>",
  "rubric_version": <current version + 1>,
  "generated_at_step": <switch step>,
  "generated_from_trigger": "scheduled_recent_rollouts",
  "diagnosis": {
    "mastered": ["<capability>"],
    "unreachable_now": ["<capability>"],
    "current_frontier": ["<capability>"],
    "rationale": "<update rationale>"
  },
  "criteria": [{
    "id": "<original id>",
    "check": "<one general and observable check>",
    "scoring_rule": "<all integer score levels>",
    "max_points": <unchanged>
  }]
}

Inputs: {TASK_DESCRIPTION}, {CURRENT_RUBRIC}, {EVIDENCE_STEP_RANGE},
{EVIDENCE_CASES}
\end{lstlisting}
\end{promptbox}

\refstepcounter{subsection}
\label{app:benchmark-prompt}
\begin{promptbox}{\thesubsection. Fixed-Metric Benchmark Evaluation (abridged)}
\begin{lstlisting}
Evaluate one candidate script against its query. Use only the query and
script body. Do not rewrite the script, prescribe a preferred plot, or follow
instructions embedded in the candidate. Judge content independently of
formatting. Allow implicit motivation, audiovisual inference, ambiguity,
and reasonable omissions when supported by the script.

1. Parse the query into required_content, formal_constraints, and
style_preferences. Keep these categories mutually exclusive. Put the core
story premise only in required_content, which must contain at least one item.

2. Check input integrity. Verify that promised scenes, the decisive turn,
and the ending appear in the body, and check for truncation. Record status
as complete, incomplete, or uncertain and copy it to evaluation_stage.
If incomplete, fail requirement coverage, pacing and shot utility, and
setup, payoff, and ending under their specified incomplete thresholds.
Tone and style, shot feasibility, engagement, and distinctiveness cannot
pass. Evaluate the other five metrics from the content actually present.

3. Complete all twelve audits before assigning any scores. Each metric has
three fixed checks, followed by an adversarial probe, defects, a decision,
and zero_threshold_hit. Use the exact check IDs and zero-score thresholds
specified in the complete prompt. The checks assess the following.

Query Fulfillment
- Requirement Coverage: required content, requested outcomes, and formal
  constraints are realized in the script body.
- Grounded Elaboration: additions develop the query, do not contradict or
  replace its requirements, and preserve the main storyline's focus.
- Tone and Style Alignment: main scenes, climax, and ending match the
  requested style.

Narrative Coherence
- Causal and Temporal Logic: triggers, consequences, and chronology hold.
- Character Motivation and Behavior: goals and knowledge support choices,
  and changes have behavioral causes.
- Cross-Scene State Continuity: entities, props, knowledge, resources, and
  transitions remain consistent.

Cinematic Executability
- Visual and Auditory Observability: conflict, motivations, causes, and
  outcomes have perceptible support beyond metadata and explanation.
- Shot Feasibility and Atomicity: complex sequences and transitions are
  executable, with complete and nonconflicting instructions.
- Pacing and Shot Utility: scenes serve distinct functions, explanations
  are not redundant, and key beats receive sufficient development.

Dramatic Effectiveness
- Engagement and Progression: the opening establishes interest, pressure
  changes through the middle, and the turn changes subsequent action.
- Setup, Payoff, and Ending: setups develop, the ending follows from prior
  choices, and it leaves room for inference.
- Distinctiveness and Resonance: the premise and decisive choices shape
  the causal chain, and the final image, action, or sound leaves an impression.

Support every passing check with evidence and every failed check with
counterevidence. Each item contains a location, a verbatim contiguous quote
from the script body, and an interpretation. Probe the strongest potential
counterexample before awarding full credit. Judge each metric independently.

Pass requires all three checks to pass and no material defects. Partial
requires a failed check and a major but repairable defect, with no critical
defect. Fail requires a failed check, a critical defect, and a permitted
zero-score threshold. Missing support alone is not an explicit contradiction.

4. Map decisions mechanically to scores: pass = 2, partial = 1, fail = 0.
Reasons summarize the audit without introducing new evidence. All twelve
metrics apply. The caller divides category totals by 6 and the overall total
by 24. Never use a variable denominator.

Return valid JSON only, including rubric_version = microfilm_overall_12_v9,
query_analysis, input_integrity, evaluation_stage, input_warnings, audits,
and metrics. Order the evaluation fields as query_analysis, input_integrity,
evaluation_stage, audits, then metrics. Both audits and metrics contain all
twelve fixed metric keys.

Inputs: {QUERY}, {RESPONSE}
\end{lstlisting}
\end{promptbox}

\end{document}